\documentclass[10pt,journal,compsoc]{IEEEtran}
\ifCLASSOPTIONcompsoc
  \usepackage[nocompress]{cite}
\else
  \usepackage{cite}
\fi
\usepackage{setspace}
\usepackage[OT2,T2A,T1]{fontenc}
\usepackage[utf8]{inputenc}
\usepackage{bm}
\usepackage[normalem]{ulem}
\usepackage{mathrsfs}

\ifCLASSINFOpdf
  \usepackage[pdftex]{graphicx}
\else
\fi
\usepackage{url}

\usepackage{rotating}
\usepackage{xcolor,colortbl}
\usepackage[pstarrows]{pict2e}

\usepackage[russian,french,english]{babel}
\usepackage{amsmath,amssymb}
\newtheorem{theorem}{Theorem}

\newcommand{\textlat}{}
\newcommand{\BC}[5]{\put(#1,#2){\resizebox{#3\unitlength}{#4\unitlength}{#5}}}
\newcommand{\BCR}[5]{\put(-#2,#1){\begin{rotate}{90}\resizebox{#3\unitlength}{#4\unitlength}{#5}\end{rotate}}}

\DeclareRobustCommand{\cyrins}[1]{%
  \begingroup\fontencoding{T2A}\fontfamily{erewhon-TLF}\selectfont%
  \foreignlanguage{russian}{#1}%
  \endgroup
}

\newcommand{\ru}[1]{\begin{otherlanguage*}{russian}{\cyrins{#1}}\end{otherlanguage*}}

\DeclareMathOperator{\D}{d}

\usepackage{tikz}
\tikzset{>=stealth}
\usepackage{pgfplots}

\pgfplotsset{scaled y ticks=false}
\usepackage{extarrows}

\begin{document}

\renewcommand{\figurename}{Fig.}

%
\title{A mathematical model for universal semantics}
%
%
%
%
\author{Weinan~E
        and~Yajun~Zhou
\IEEEcompsocitemizethanks{\IEEEcompsocthanksitem Weinan E is with Department of Mathematics and Program in Applied and Computational Mathematics, Princeton University, and Beijing Institute of Big Data Research\protect\\
E-mail: \protect\url{weinan@math.princeton.edu}\IEEEcompsocthanksitem Yajun Zhou is with  Beijing Institute of Big Data Research\protect\\Email. \protect\url{yajun.zhou.1982@pku.edu.cn}}
\thanks{}}

\IEEEtitleabstractindextext{%
\begin{abstract}
We  characterize the meaning of
words with language-independent numerical fingerprints, through a mathematical analysis of recurring patterns in texts. Approximating texts by Markov processes on a long-range time scale, we are able to extract topics, discover synonyms, and sketch semantic fields from  a particular document of moderate length, without consulting external knowledge-base or thesaurus. Our Markov semantic model allows us to  represent each topical concept by a low-dimensional vector, interpretable as algebraic invariants in succinct statistical operations on the document, targeting local environments of individual words. These language-independent semantic representations enable a robot reader to both understand short texts in a given language (automated question-answering) and match medium-length texts across different languages (automated word translation). Our  semantic fingerprints quantify local meaning of words in 14 representative languages across 5 major language families, suggesting a universal and cost-effective mechanism by which  human languages are processed at the semantic level. Our protocols and source codes are publicly available on \url{https://github.com/yajun-zhou/linguae-naturalis-principia-mathematica}\end{abstract}

}

\maketitle

\IEEEdisplaynontitleabstractindextext

%
\IEEEpeerreviewmaketitle

\IEEEraisesectionheading{\section{Introduction}\label{sec:introduction}}

%
%
%
%
\IEEEPARstart{A}{ quantitative} model for the meaning of words not only helps us understand how we transmit information and absorb knowledge, but also provides foothold for algorithms in machine processing of natural language texts. Ideally,  a universal mechanism of semantics should be based  on numerical characteristics of human languages,  transcending concrete written and spoken forms of verbal messages.        In this work, we  demonstrate,  in both theory and practice,  that the time structure  of recurring language patterns is a good candidate for such a universal  semantic mechanism. Through  statistical analysis of   recurrence times and hitting times, we numerically characterize   connectivity and association of individual concepts, thereby devising language-independent semantic fingerprints (LISF).

Concretely speaking, we define semantics through algebraic invariants of a stochastic text model that approximately governs the empirical hopping rates on a web of word patterns. Such a stochastic model explains the distribution of recurrence times and outputs recurrence eigenvalues as semantic fingerprints. Statistics of recurrence times allow machines to tell non-topical words from topical ones. A comparison of hitting  and recurrence times further generates quantitative fingerprints for topics, enabling machines to overcome language barriers in translation tasks and perform associative reasoning in comprehension tasks,       like humans.

Akin to the physical world, there is a hierarchy of length scales in languages.  On short scales such as syllables, words, and phrases,  human languages do not  exhibit a universal pattern related to semantics. Except for a few onomatopoeias, the sounds of words do not affect their meaning \cite{Saussure1949}. Neither do morphological parameters \cite{Pinker1988} (say, singular/plural, present/past) or syntactic r\^oles \cite{Chomsky2002syntactic} (say, subject/object, active/passive).  In short, there are no universal semantic mechanisms at the phonological, lexical or syntactical levels \cite{Friederici1999}. Grammatical ``rules and principles'' \cite{Pinker1988,Chomsky2002syntactic}, however typologically diverse,  play no definitive r\^ole in determining the  inherent meaning of a word.

\begin{figure*}\begin{center}\begin{footnotesize}$\mathsf W_i:=\,$happ(ier$|$ily$|$iness$|$y)$\,\equiv\,$\{\textit{happier}, \textit{happily}, \textit{happiness}, \textit{happy}\},\quad $\mathsf W_j:=\,$marr(iage$|$ied$|$y)$\,\equiv\,$\{\textit{marriage}, \textit{married},  \textit{marry}\}
\end{footnotesize}\end{center}
\vspace{-1em}

\begin{picture}(500,25)(-5,-8)
\setlength{\unitlength}{1pt}
\put(3,6){\fontsize{0.1865cm}{1em}\selectfont{\texttt{... LOREM IPSUM \textcolor{magenta}{HAPPY}\uline{ DOLOR SIT AMET, }\textcolor{magenta}{HAPPY}\uline{, CONSECTETUR ADIPISCING UNHAPPY ELIT, }\textcolor{magenta}{HAPPINESS}\uline{ SED }\textcolor{magenta}{HAPPY}\uline{ DO }\textcolor{magenta}{HAPPY}\uline{  EIUSMOD TEMPOR }\textcolor{magenta}{HAPPIER}, INCIDIDUNT UT LABORE ...}}}
\put(3,0){\fontsize{0.1865cm}{1em}\selectfont{\texttt{\phantom{... LOREM IPSUM \textcolor{magenta}{HAPPY}}{\textcolor[rgb]{0.75,.75,.75}{HAPPINESS}\phantom{T AMET, }}\textcolor{magenta}{\phantom{HAPPY}}{\textcolor[rgb]{0.75,.75,.75}{HAPPINESS}\phantom{ETUR ADIPISCING UNHAPPY ELIT, HAPPINESS SED HAPPY DO HAPPY}\textcolor[rgb]{0.75,0.75,0.75}{HAPPINESS}}}}}
\textcolor{blue}{\put(99,4){\line(0,-1){10}}
\put(107,-0.5){\vector(-1,0){8}}
\put(124,4){\line(0,-1){10}}
\put(106,-0.5){\vector(1,0){18}}
\put(107,-7){\fontsize{0.12cm}{1em}\selectfont{$L_{ii}$}}
\put(169,4){\line(0,-1){10}}
\put(208,-0.5){\vector(-1,0){39}}
\put(264.5,4){\line(0,-1){10}}
\put(212,-7){\fontsize{0.12cm}{1em}\selectfont{$L_{ii}$}}
\put(205.5,-0.5){\vector(1,0){59}}
\put(389,-0.5){\vector(-1,0){7}}
\put(382,4){\line(0,-1){10}}
\put(387.5,-0.5){\vector(1,0){17}}
\put(388,-7){\fontsize{0.12cm}{1em}\selectfont{$L_{ii}$}}
\put(404.5,4){\line(0,-1){10}}
}
\end{picture}

\vspace{-.25em}
\begin{picture}(500,50)(-5,-35)
\setlength{\unitlength}{1pt}
\put(3,6){\fontsize{0.1865cm}{1em}\selectfont{\texttt{... LOREM IPSUM,  \textcolor{red}{MARRIAGE}{ DOLOR SIT AMET, }\textcolor{magenta}{HAPPY}\uline{, CONSECTETUR ADIPISCING }\textcolor{red}{MARRIED} ELIT, \textcolor{red}{MARRY}{ SED }\textcolor{magenta}{HAPPILY}{ DO }\textcolor{magenta}{HAPPILY}\uline{  EIUSMOD TEMPOR }\textcolor{red}{MARRIED} INCIDIDUNT UT\ LAB ...}}}
\put(3,0){\fontsize{0.1865cm}{1em}\selectfont{\texttt{\phantom{... LOREM IPSUM HAPPINESS DOLOR SIT AMET, HAPPY}\textcolor[rgb]{0.75,0.75,0.75}{HAPPINESS}\phantom{ETUR ADIPISCING UNHAPPY ELIT, UNHAPPY SED HAPPIER DO HAPPY}\textcolor[rgb]{0.75,0.75,0.75}{HAPPINESS}}}}
\textcolor{blue}{\put(182,4){\line(0,-1){10}}
\put(233,4){\line(0,-1){10}}
\put(198,-0.5){\vector(-1,0){16}}
\put(203,-7){\fontsize{0.12cm}{1em}\selectfont{$L_{ij}$}}
\put(196.5,-0.5){\vector(1,0){36}}
\put(395,4){\line(0,-1){10}}
\put(417,4){\line(0,-1){10}}
\put(408,-0.5){\vector(-1,0){13}}
\put(400,-7){\fontsize{0.12cm}{1em}\selectfont{$L_{ij}$}}
\put(400.5,-0.5){\vector(1,0){16}}}
\put(3,-12){\fontsize{0.1865cm}{1em}\selectfont{\texttt{\phantom{... LOREM IPSUM HAPPINESS DOLOR SIT AMET, HAPPY}\uline{\phantom{, CONSECTETUR ADIPISCING UNHAPPY ELIT, }}}}}
\put(3,-18){\fontsize{0.1865cm}{1em}\selectfont{\texttt{\phantom{... LOREM IPSUM HAPPINESS DOLOR SIT AMET, HAPPY}\textcolor[rgb]{0.75,0.75,0.75}{HAPPINESS}}}}\textcolor{blue}{\put(182,-14){\line(0,-1){10}}
\put(277,-14){\line(0,-1){10}}
\put(218,-18.5){\vector(-1,0){36}}
\put(225,-25){\fontsize{0.12cm}{1em}\selectfont{$L_{ij}$}}
\put(210.5,-18.5){\vector(1,0){66}}
}
\end{picture}\vspace{-2em}
\caption{Counting long-range transitions between word patterns. A transition from $ \mathsf W_i$ to $\mathsf W_j$ counts towards long-range statistics, if the \textit{underlined} text fragment in between contains no occurrences of $ \mathsf W_i$, and lasts strictly longer than the longest word in $ \mathsf W_i\cup\mathsf W_j$. For each long-range transition, the effective fragment length $ L_{ij}$  discounts the length of the longest word in  $ \mathsf W_i\cup\mathsf W_j$.  \label{fig:NijLij}}\end{figure*}
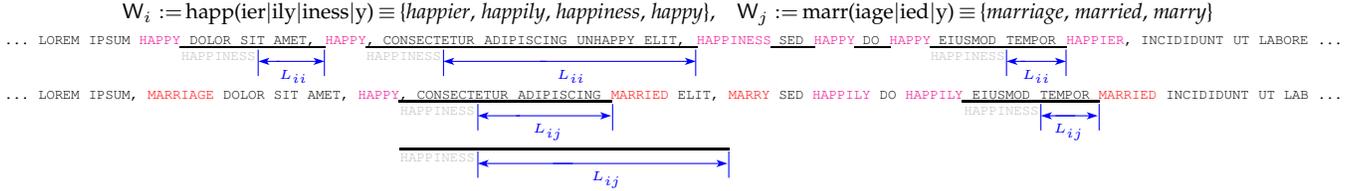

Motivated by the observations above, we will build our quantitative semantic model on long-range and language-independent textual features.  Specifically, we will measure the    lengths  of text fragments flanked by   \textit{word patterns} of interest   (Fig.~\ref{fig:NijLij}). Here, a word pattern is a collection of \textit{content words} that are identical up to morphological parameters and syntactic r\^oles. A content word signifies definitive concepts (like  \textit{apple}, \textit{eat}, \textit{red}), instead of serving purely grammatical or logical functions (like \textit{but}, \textit{of}, \textit{the}). Fragment length statistics will tell us how tightly/loosely one concept is connected to another. This in turn, will provide us with  quantitative criteria for inclusion/exclusion of different concepts within the same (computationally constructed) semantic field. Such statistical semantic  mining will then pave the way for  machine comprehension and machine translation.\section{Methodology}

We quantify the time structure of an individual  word pattern $ \mathsf W_i$ through the statistics of its recurrence times $ \tau_{ii}$. We characterize the dynamic impact of  a word pattern $ \mathsf W_i$  on another word pattern $ \mathsf W_j$ by the statistics of their hitting times $ \tau_{ij}$. In what follows, we will describe the statistical analyses of  $ \tau_{ii}$ and
$ \tau_{ij}$, on which we build a language-independent Markov model for semantics.

\begin{figure*}
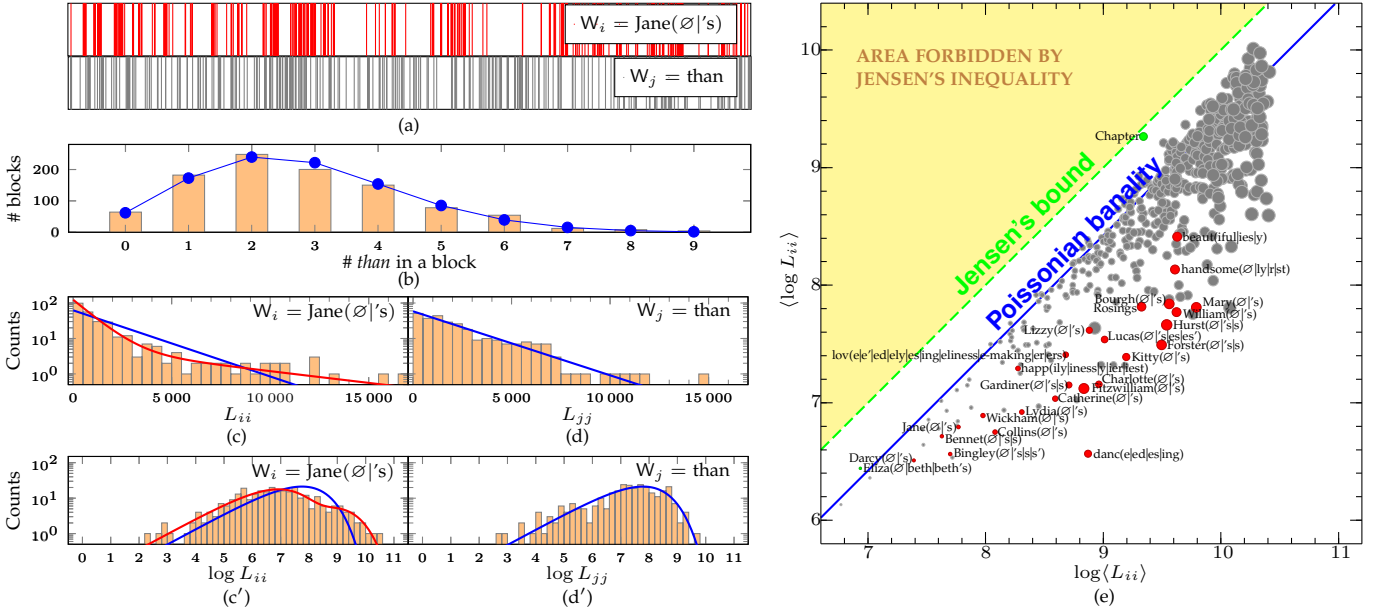


\vspace{-1.75em}

\begin{minipage}{0.6\textwidth}\begin{center}
\hspace{0em}

\end{minipage}\vspace{-1.25em}
\caption{Statistical analysis of recurrence times and topicality. (a)~Barcode representations  (adapted from~\cite[Fig.~2]{HerreraPury2008}) for the coverage of $\mathsf W_i=\text{Jane}(\varnothing|\text{'s})$ (291 occurrences) and $ \mathsf W_j=\text{than}$ (282 occurrences) in the whole text of \textit{Pride and Prejudice}. Horizontal axis scales linearly with respect to the text length  measured in the number of constituting letters, spaces and punctuation marks. (b)~Counts  of the word  \textit{than} within a  consecutive block of 1217 words (spanning about 1\% of  the entire text), drawn from 1000 randomly chosen  blocks, fitted to a Poisson distribution with mean 2.776 (\textit{blue} curve). (c)~Histogram of effective fragment length $ L_{ii}$ (see Fig.~\ref{fig:NijLij} for its definition) for the topical pattern $\mathsf W_i=\text{Jane}(\varnothing|\text{'s})$, fitted to an exponential distribution (\textit{blue} line in the semi-log plot) and a weighted mixture of two exponential distributions $ c_1 k_1e^{-k_1t}+c_{2}k_2e^{-k_2t}$ (\textit{red} curve, with $ c_1:c_{2}\approx1:3$, $ k_1:k_2\approx1:7$). (d)~Histogram of  $ L_{jj}$ for the function word $ \mathsf W_j=\text{than}$, fitted to an exponential distribution (\textit{blue} line in the semi-log plot).   All the parameter estimators in panels b--d are based on maximum likelihood. (c$ '$)--(d$'$)~Reinterpretations of panels  c--d, with logarithmic binning on the horizontal axes, to give  fuller coverage of the dynamic ranges for the statistics. (e)~Recurrence statistics for word patterns in Jane Austen's  \textit{Pride and Prejudice},  where $\langle\cdots\rangle$ denotes averages over $ n_{ii}$ samples of long-range transitions. Data points in \textit{gray}, \textit{green} and \textit{red} have radii $ \frac{1}{4\sqrt{\smash[b]{n_{ii}}}}$. Labels for proper names and some literary motifs are attached next to the corresponding colored dots.  Jensen's bound (\textit{green dashed line}) has  unit slope and zero intercept. Exponentially distributed recurrence statistics reside on  the line of Poissonian banality (\textit{blue line}), with  unit slope and negative intercept. \textit{Red} (resp.~\textit{green}) dots mark significant downward (resp.~upward) departure from the \textit{blue line}.\label{fig:rec_topic}}\end{figure*}

\subsection{Recurrence times and topicality\label{subsec:rec_topic}}Assuming  uniform
reading speed,\footnote{On the scale of words (rather than phonemes), this assumption works fine in most  languages that are written alphabetically. However, this working hypothesis   does not extend to Japanese texts, which  interlace Japanese syllabograms (lasting one mora per written unit) with Chinese ideograms (lasting one or more morae per  written unit).} we measure   the  recurrence times  $ \tau_{ii}$ for a word pattern $ \mathsf W_i$ through   $ n_{ii}$ samples of the effective fragment lengths $ L_{ii}$ (Figs.~\ref{fig:NijLij}, \ref{fig:rec_topic}a).
Here, while counting as in Fig.~\ref{fig:NijLij},  we
ignore
contacts between short-range neighbors, which may involve language-dependent redundancies.\footnote{For example, a German phrase \textit{liebe Studentinnen und Studenten }with short-range recurrence is the gender-inclusive equivalent of the English expression \textit{dear students}. Some Austronesian languages (such as Malay and Hawaiian) use reduplication for  plurality or emphasis.}

\subsubsection{Recurrence of non-topical patterns}
 In a memoryless (hence banal) Poisson process (Fig.~\ref{fig:rec_topic}b), recurrence times are exponentially distributed (Fig.~\ref{fig:rec_topic}d,d$'$). The same is also true for word recurrence  in a randomly reshuffled text \cite{HerreraPury2008}. If  we have $n_{ii}$ independent samples of exponentially distributed random variables $L_{ii} $, then the statistic  $ \delta_{i}:=\log\langle L_{ii}\rangle-\langle\log L_{ii}\rangle- \gamma_0+\frac{1}{2n_{ii}}$ satisfies an inequality\begin{align}\left|\delta_{i} \right|<\frac{2}{\sqrt {\smash[b]{n_{ii}}}}\sqrt{\frac{\pi^2}{6}-1-\frac{1}{2n_{ii}}}\label{ineq:95_conf_large_N}\end{align}with probability 95\% (see Theorem~\ref{thm:YN_dist} in Appendix \ref{app:two-sigma_Poisson} for a two-sigma rule).  Here, $  \gamma_0:=\lim_{n\to\infty}\left(-\log n+\sum_{m=1}^n\frac1m\right)$ is the Euler--Mascheroni constant.

As a working definition, we consider a word pattern
$ \mathsf W_i$
\textit{non-topical} if  its $ n_{ii}$ counts  of  effective fragment lengths $ L_{ii}$ are exponentially distributed $ \mathbb P(L_{ii}>t)\sim e^{-kt}$, within  95\% margins of error [that is, satisfying \eqref{ineq:95_conf_large_N} above].

\begin{figure*}[t]
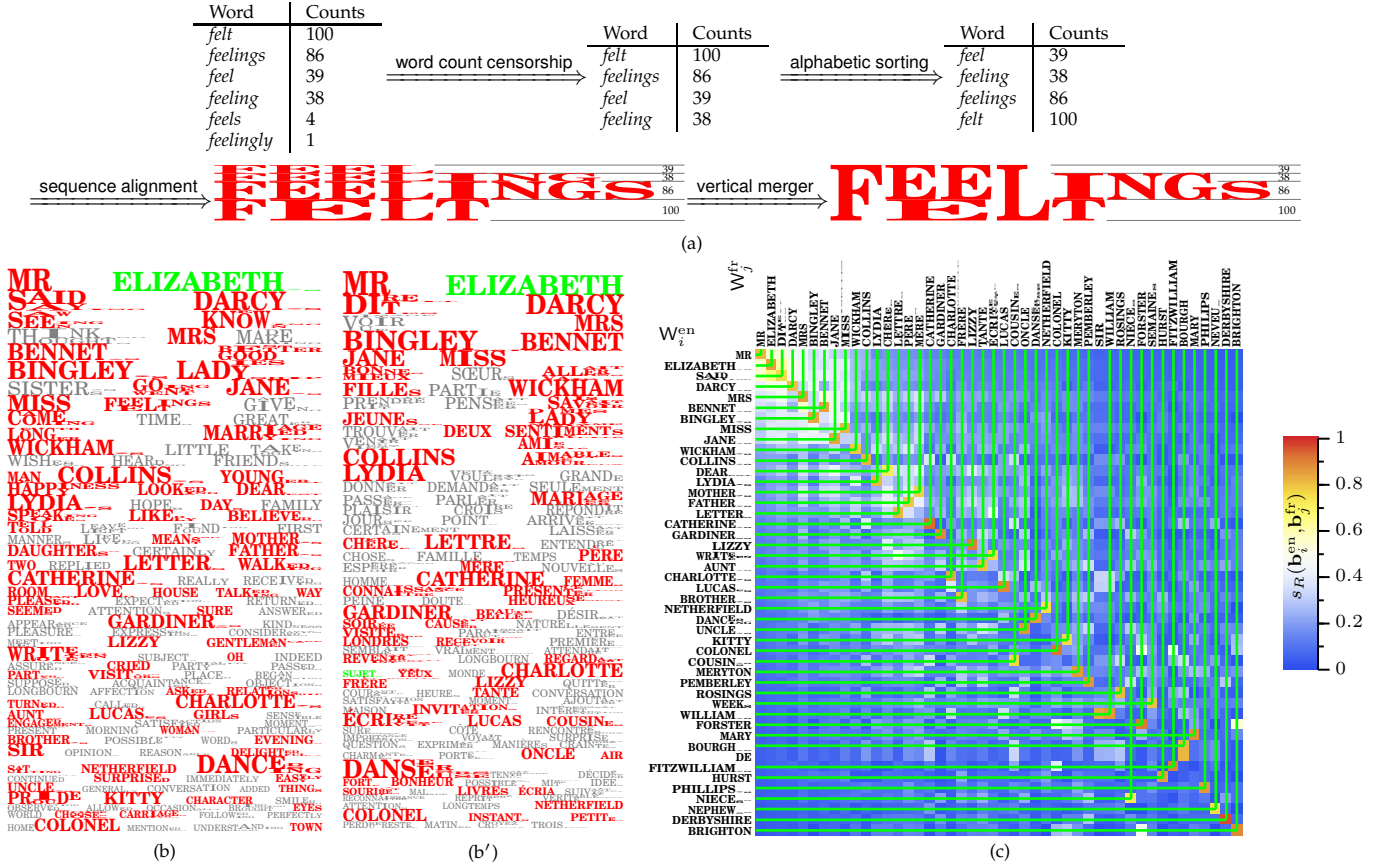


\begin{minipage}{1\textwidth}\begin{center}
\begin{minipage}{.14\textwidth}
\begin{scriptsize}
\begin{tabular}{l|l}Word&Counts\\\hline
\textit{felt}&100\\
\textit{feelings}&86\\
\textit{feel}&39\\
\textit{feeling}&38\\
\textit{feels}&4\\
\textit{feelingly}&1
\end{tabular}\end{scriptsize}\end{minipage}$\xLongrightarrow{\text{\textsf{word count censorship} }}$\begin{minipage}{.14\textwidth}
\begin{scriptsize}%

\end{minipage}

\vspace{.5em}
\begin{minipage}{.23\textwidth}\begin{center}{\scriptsize (b)}\end{center}\end{minipage}\hspace{.015\textwidth}\begin{minipage}{.205\textwidth}\begin{center}{\scriptsize (b$'$)}\end{center}\end{minipage}\begin{minipage}{.55\textwidth}\begin{center}{\scriptsize (c)}\end{center}\end{minipage}

\vspace{-1em}
\caption{Automated topic extraction and raw alignment across bilingual corpora. (a)~Schematic diagram illustrating our graphical representation of morphologically related words (identified by supervised algorithms in  \textit{Supplementary Materials}) in a word pattern.  To avoid unprintably small characters, rarely occurring forms (less than 5\% of the total sum of all the words ranked above) are ignored in graphical display. To enhance the visibility of word stems, we print shared letters only once, and compress other letters vertically, with heights proportional to their corresponding word counts. (b)~Word patterns  $ \mathsf W_i$     in Jane Austen's \textit{Pride and Prejudice}, sorted by descending $ n_{ii}$, with font size    proportional to the square root of  $ {e^{-\langle \log L_{ii}\rangle}}$ (a better indicator of reader's impression than the number of recurrences $ n_{ii}\propto e^{-\log\langle L_{ii}\rangle}$).    Topical (that is, significantly non-Poissonian) patterns painted in  \textit{red} (resp.~\textit{green}) reside below (resp.~above) the critical line of Poissonian banality (\textit{blue line} in Fig.~\ref{fig:rec_topic}e), where the deviations exceed the error margin prescribed in  \eqref{ineq:95_conf_large_N} of \S\ref{subsec:rec_topic}. (b$'$)~A similar service on a French version of \textit{Pride and Prejudice} (tr.~Valentine Leconte
\& Charlotte Pressoir). (c)~A low-cost and low-yield word translation, based on chapter-wise word counts $ \mathbf b_i^{\mathrm{en}}$ and $\mathbf b_j^{\mathrm{fr}}$. Ru\v zi\v cka similarities $ s_{R}(\mathbf b_i^{\mathrm{en}},\mathbf b_j^{\mathrm{fr}})$ between selected topics (sorted by descending $ n_{ii}\geq20$) in English and French versions  of \textit{Pride and Prejudice}.  Rows and columns with maximal  $ s_{R}(\mathbf b_i^{\mathrm{en}},\mathbf b_j^{\mathrm{fr}})$  less than $0.7$ are not shown. Correct matchings are indicated by green cross-hairs.\label{fig:topic_ext}}
\end{figure*}

\subsubsection{Recurrence of topical patterns}
In contrast, we consider a word pattern $ \mathsf W_i$ \textit{topical} if its diagonal statistics $ n_{ii},L_{ii}$ constitute significant departure   from the Poissonian line $ \langle\log L_{ii}\rangle-\log\langle L_{ii}\rangle+\gamma_0=0$ (Fig.~\ref{fig:rec_topic}e, blue line), violating the bound in \eqref{ineq:95_conf_large_N}.

Notably,  most data points for  topics  (colored dots on Fig.~\ref{fig:rec_topic}e) in Jane Austen's \textit{Pride and Prejudice} mark systematic downward departures from the Poissonian line. This suggests that the topical recurrence times  $ \tau=L_{ii}$   follow weighted mixtures of exponential distributions (Fig.~\ref{fig:rec_topic}c,c$'$):\begin{align}&\mathbb P(\tau>t)\sim\sum_mc_me^{-k_mt},\label{eq:multi_exp}
\end{align}$\left(\text{where }c_m,k_m>0,\text{ and }\sum_mc_m=1\right)$, which impose an inequality constraint on the   recurrence time $ \tau=L_{ii}$:\begin{align}&\langle\log L_{ii}\rangle-\log\langle L_{ii}\rangle+\gamma_0\notag\\={}& \sum_m c_m\log\frac{1}{k_m}-\log\sum_m \frac{c_m}{k_m}\leq 0.\label{ineq:log_Lf}
\end{align}

\subsubsection{Raw alignment of topical patterns}If a word pattern $ \mathsf W_i$ qualifies as a topic  by our definition (Fig.~\ref{fig:topic_ext}b,b$'$), then the signals in its coarse-grained timecourse (say, a vector  $\mathbf b_i=(b_{i,1},\dots,b_{i,61}) $ representing   word counts in each chapter of \textit{Pride and Prejudice}) are not overwhelmed by Poisson  noise.

This vectorization scheme, together with the Ru\v zi\v cka similarity \cite{Ruzicka1958} \begin{align} s_R(\mathbf b_i^{\mathrm A},\mathbf b_j^{\mathrm B}):=\frac{\Vert \smash[b]{\mathbf b_i^{\mathrm A}\wedge\mathbf b_j^{\mathrm B}}\Vert_1}{\Vert \smash[b]{\mathbf b_i^{\mathrm A}\vee\mathbf b_j^{\mathrm B}}\Vert_{1}}\end{align} between two vectors with non-negative entries, allow us to  align some topics found in parallel versions of the same document, in languages A and B
 (Fig.~\ref{fig:topic_ext}c). Here, in the definition of  the Ru\v zi\v cka similarity,   $ \wedge$ (resp.~$ \vee$) denotes component-wise minimum (resp.~maximum) of vectors; $ \Vert\mathbf b\Vert_1$ sums over all the components in $ \mathbf b$.

\subsection{Markov text model}
\subsubsection{Transition probabilities via pattern analysis}The diagonal statistics $ n_{ii},L_{ii}$ (Fig.~\ref{fig:NijLij}) have enabled us to extract topics automatically through recurrence time analysis (Figs.~\ref{fig:rec_topic}e and \ref{fig:topic_ext}b,b$'$). The off-diagonal statistics  $ n_{ij},L_{ij}$ (Fig.~\ref{fig:NijLij})  will allow us to determine how strongly one word pattern $ \mathsf W_i$ binds to another word pattern $\mathsf W_j $, through hitting time analysis. In an empirical Markov matrix $  \mathbf{ P}=( p_{ij})$,  the long-range transition rate $ p_{ij}$ is estimated by \begin{align} p_{ij}:=\frac{n_{ij}e^{-\langle\log L_{ij}\rangle}}{\displaystyle\sum_{k=1}^Nn_{ik}e^{-\langle\log L_{ik}\rangle}},\label{eq:pij_est}
\end{align}where $ n_{ij}$ counts the number of long-range transitions from $ \mathsf  W_i$ to $\mathsf  W_j $, and $ L_{ij}$ is a statistic that measures the effective fragment lengths of such transitions (Fig.~\ref{fig:NijLij}).

\subsubsection{Equilibrium state and detailed balance}Numerically,  we find that  our empirical Markov matrix $  \mathbf{ P}=( p_{ij})$ defined in \eqref{eq:pij_est} is a fair approximation to an \textit{ergodic}\footnote{If a Markov chain is  ergodic,  then there is a strictly positive probability to transition from any Markov state (that is, any individual word pattern in our model) to any other state, after finitely many steps.} matrix $ \mathbf P^{*}=( p_{ij}^{*})$, which in turn, governs the stochastic hoppings between content word patterns  during text generation.
\begin{figure}
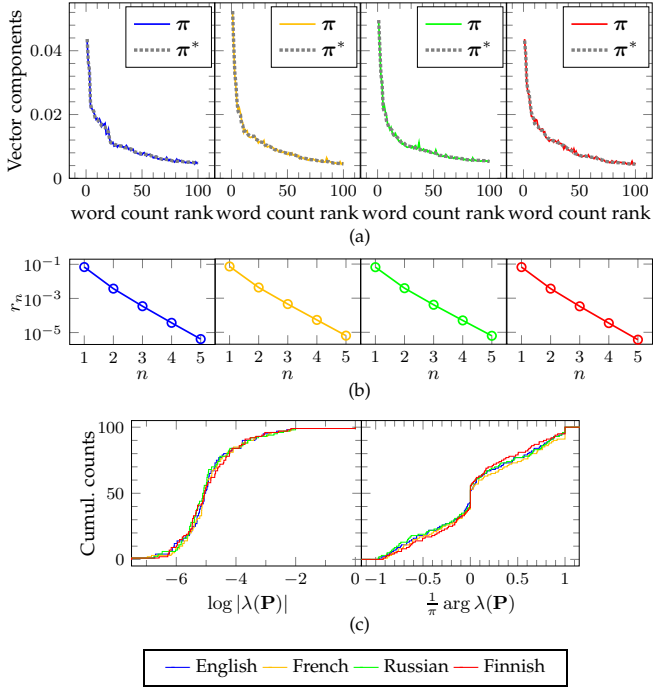

\vspace{2em}

\hspace{-2.39em}\begin{minipage}{0.125\textwidth}
\vspace{-1.78em}

\vspace{-1.5em}
\begin{center}{\scriptsize\hspace{.6em} (c)}\end{center}
\vspace{-.2cm}
\begin{center}
\hspace{.3em}\scriptsize\fbox{\;\;\,\textcolor{blue}{--\!---} English \textcolor{orange!50!yellow}{--\!---} French \textcolor{green}{--\!---} Russian  \textcolor{red}{--\!---} Finnish\;\;\,}
\end{center}

\vspace{-1em}\caption{Quantitative properties of Markov text model. (a)~Dominant eigenvector $ \bm{ \pi}$  of a $ 100\times100$ Markov matrix  $ \mathbf{ P} $, computed  from one of the four versions of  \textit{Pride and Prejudice}, in comparison with $ \bm \pi^{*}$,  the list of normalized frequencies for top 100 word patterns. (b)~Precipitous decays of $ r_n:=\frac12\sum_{1\leq i,j\leq 100}\big| \pi_i p_{ij}^{(n)}- \pi_j p_{ji}^{(n)}\big|$ from the initial value $ r_1\approx0.07$, for matrix powers ${ \mathbf P}^n=( p_{ij}^{(n)})_{1\leq i,j\leq 100}$ constructed from four versions of  \textit{Pride and Prejudice}. (In contrast, one has  $ r_1\approx0.33$ for a random  $100\times100$ Markov matrix.) Such quick relaxations support our working hypothesis about detailed balance $ \pi_i ^{*}p_{ij}^{*}=\pi_j^{*}p_{ji}^*$. (c)~Distributions of eigenvalues $ \lambda$ of empirical Markov matrices $ \mathbf{ P}$, with  nearly language-independent   modulus $ |\lambda(\mathbf{ P})|$ and  phase-angle $ \arg\lambda( \mathbf{ P})$.\label{fig:Markov_model}}\end{figure}

Each ergodic Markov matrix  $  \mathbf{ P}^{*}=( p_{ij}^{*})_{1\leq i,j\leq N}$    possesses a unique equilibrium state $ \bm \pi^{*}=(\pi_{i}^{*})_{1\leq i\leq N}$. The equilibrium state $ \bm\pi^{*}$ represents a probability distribution (that is, $ \pi_i^{*}\geq0$ for $ 1\leq i\leq N$ and $ \sum_{i=1}^N\pi_i^{*}=1$) that satisfies $ \bm \pi^{*} \mathbf P^{*}=\bm \pi^{*}$ (that is, $ \sum _{1\leq i\leq N}\pi_i ^{*}p_{ij}^{*}=\pi_j^{*}$ for  $ 1\leq j\leq N$). In our numerical experiments, the dominant eigenvector $ \bm \pi$ (satisfying $ \bm \pi \mathbf P=\bm \pi$) consistently reproduces word frequency statistics that are proportional to the  ideal equilibrium state $ \bm \pi^*$ (Fig.~\ref{fig:Markov_model}a).

Furthermore, through numerical experimentation, we find that our empirical Markov matrix $ \mathbf P\approx\mathbf P^*$ approximately honors the detailed balance condition  $ \pi_i^{*}p_{ij}^*= \pi_j^*p_{ji}^*$ for  $ 1\leq i,j\leq N$. The approximation    $ \pi_ip_{ij}^{(n)}\approx \pi_jp_{ji}^{(n)}$  becomes closer as we go to  higher iterates $ \mathbf P^n=(p_{ij}^{(n)})$, where $n$ is a small positive integer (Fig.~\ref{fig:Markov_model}b).

On an ergodic Markov chain with detailed balance, one can show that recurrence times are distributed as  weighted mixtures of exponential decays (see Theorem~\ref{thm:cm_R} in Appendix~\ref{app:rec_time}), thus offering a theoretical explanation for    \eqref{eq:multi_exp}.
 \subsubsection{Spectral invariance under translation}
The spectrum
$\sigma( \mathbf{ P})$
(collection of eigenvalues) is approximately invariant against translations of texts  (Fig.~\ref{fig:Markov_model}c), which can be explained by   a matrix equation   \begin{align}
 \mathbf P_{\mathrm A}\mathbf T_{\mathrm A\to\mathrm B}=\mathbf T_{\mathrm A\to\mathrm B}\mathbf P_{\mathrm B}.\label{eq:PTTP}\end{align}Here, both sides of the identity  above quantify the transition probabilities from words in language A to words in language B, from the impressions of   Alice and Bob, two monolingual readers in a thought experiment.   On the left-hand side, Alice first processes the input in her native language A by a Markov matrix $ \mathbf P_{\mathrm A}$, and then translates into language B, using a dictionary matrix $ \mathbf T_{\mathrm A\to\mathrm B}$; on the right-hand side, Bob needs to first translate the input into language B, using the same dictionary  $\mathbf  T_{\mathrm A\to\mathrm B}$, before brainstorming in his own native language, using $   \mathbf P_{\mathrm B}$. Putatively, the matrix equation holds because semantic content is shared by native speakers of different languages. In the ideal scenario where translation is lossless (with invertible  $ \mathbf T_{\mathrm A\to\mathrm B}$), the Markov matrices   $ \mathbf P_{\mathrm A}$ and  $ \mathbf P_{\mathrm B}$ are indeed linked to each other by a similarity transformation that leaves their spectrum intact.
\subsection{Localized Markov matrices and semantic cliques}
\subsubsection{Semantic contexts for recurrent  topics}
Specializing  spectral invariance  to individual {topical patterns}, we will be able to generate  semantic fingerprints through a list of topic-specific and language-independent eigenvalues. Here, we will be particularly interested in recurrence eigenvalues of individual topical patterns, which correspond to multiple decay rates in the weighted mixtures of exponential distributions.

Unlike  the single exponential decays associated to non-topical recurrence patterns, the multiple exponential decay modes  will  enable our robot reader to easily discern one topic from another.
In general, it is  numerically challenging  to  recover multiple exponential decay modes from a limited amount of  recurrence time measurements \cite{BJ2006}. However, in text processing, we can circumvent such difficulties by  off-diagonal statistics $n_{ij}$ and $L_{ij}$ that provide  semantic contexts for individual topical patterns.


To quantitatively define the semantic content of a topical pattern  $ \mathsf W_i$, we   specify a local, directed, and weighted graph, corresponding to a localized Markov transition matrix $ \mathbf{ P}^{[i]}$.
\subsubsection{Localized Markov contexts of topical patterns}

\begin{figure*}
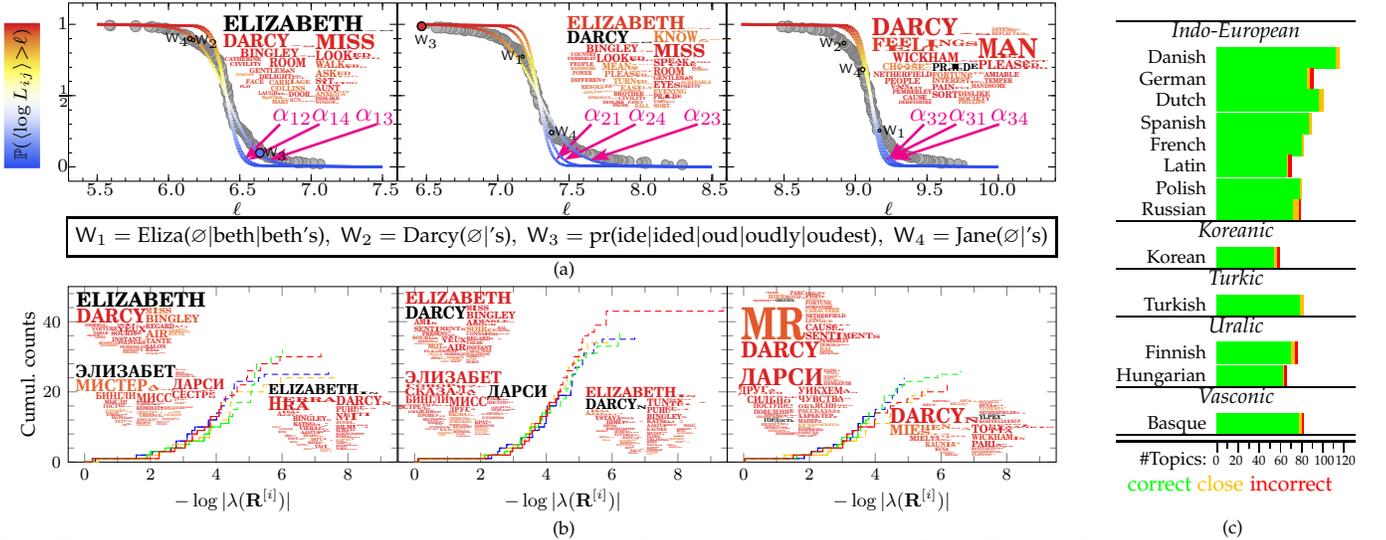


\vspace{1.5em}
\begin{minipage}{.8\textwidth}
\hspace{1em}\begin{minipage}{0.307\textwidth}
\setlength{\unitlength}{2.7pt}

\end{minipage}

\hspace{-.1em}\hspace{3em}\fbox{\footnotesize$\mathsf W_1=\text{Eliza(\ensuremath{\varnothing}\ensuremath{|}beth\ensuremath{|}beth's)},\; \mathsf W_2=\text{Darcy(\ensuremath{\varnothing}\ensuremath{|}'s)},\;\mathsf W_3=\text{pr(ide\ensuremath{|}ided\ensuremath{|}oud\ensuremath{|}oudly\ensuremath{|}oudest)},\;\mathsf W_4=\text{Jane(\ensuremath{\varnothing}\ensuremath{|}'s)}$}

\vspace{-.5em}
\begin{center}{\scriptsize\hspace{2.5em}(a)}\end{center}
\vspace{1em}
\begin{minipage}{\textwidth}\fontfamily{pnc}\selectfont\begin{spacing}{0}
}
\vspace{-1em}
\begin{center}{\footnotesize\textcolor{green}{correct} \textcolor{orange!50!yellow}{close} \textcolor{red}{incorrect} }

\vspace{0.5em}
{\scriptsize(c)}\end{center}

\end{minipage}

\vspace{-23em}

\caption{Semantic cliques and their applications to word translation. (a)~Empirical distributions of  $ \langle \log L_{ij}\rangle$  in \textit{Pride and Prejudice},  as \textit{gray} and \textit{colored} dots with radii $ \frac{1}{4\sqrt{\smash[b]{n_{ij}}}}$, compared to Gaussian model $ \alpha_{ij}(\ell)$ (\textit{colored} curves parametrized by  \eqref{eq:sigmoidal_modulation} and \eqref{eq:sigmoid_para}). The numerical samplings of  $\mathsf W_j$'s exhaust all the textual patterns  available in the novel, including topical word patterns, non-topical word patterns and function words. Only those textual patterns with over 40 occurrences are displayed as data points. \textit{Inset} of each frame shows the semantic clique $\mathscr S_i$ surrounding topic $ \mathsf W_i$ (painted in \textit{black}),  color-coded by the $ \alpha_{ij}(\langle \log L_{ij}\rangle)$ score.   The areas of the bounding boxes for individual word patterns are proportional to  the components of  $ \bm{\pi}^{[i]}$ (the equilibrium state  of $\mathbf{ P}^{[i]} $).
(b)~Distributions for the magnitudes of eigenvalues (LISF) in the recurrence matrices $ \mathbf{ R}^{[i]}$, for three concepts from four versions of  \textit{Pride and Prejudice}. The color encoding for languages follows  Fig.~\ref{fig:Markov_model}.
The largest  $ \lfloor e^{\eta_i}\rfloor$ magnitudes of eigenvalues  are displayed as \textit{solid} lines, while the remaining terms  are  shown in \textit{dashed} lines.
\textit{Inset} of each frame shows the semantic clique $\mathscr S_i$, counterclockwise from top-left, in French, Russian  and Finnish. (c)~Yields from bipartite matching of  LISF (see Fig.~\ref{fig:vec_align} for English-French) for topical words  between the English original of \textit{Pride and Prejudice} and its translations into 13  languages out of 5 language families.  \label{fig:aff}}\vspace{-1em}\end{figure*}

To localize, we need to remove edges between two vertices   $ \mathsf W_i$ and $ \mathsf W_j$,
when the hitting times  $  L_{ij}$ and $  L_{ji}$  are ``long enough'' relative to what one could na\"ively expect from recurrence time statistics $n_{ij},{n_{ji}}$ and $L_{ii}, L_{jj}$.
 Here, for na\"ive expectation, we approximate the probability $\mathbb P (\langle \log L_{ij}\rangle>\ell)$ by a Gaussian model $ \alpha_{ij}(\ell)$ (colored curves in Fig.~\ref{fig:aff}a) \begin{align}\mathbb P (\langle \log L_{ij}\rangle>\ell)\approx \alpha_{ij}(\ell):=\sqrt{\frac{n_{ij}}{{2\pi\beta_{i}}}}\int_{\ell}^\infty e^{-\frac{n_{ij}(x-\ell_i)^2}{2\beta_{i}}}\mathrm{d}\, x,\label{eq:sigmoidal_modulation}\end{align}whose mean and variance  are deducible from $ n_{ij}$ and $ L_{ii}$ (see Theorem~\ref{thm:mean_var} in Appendix~\ref{app:sigm}):\begin{align}\ell_i:=\frac{\langle L_{ii}\log  L_{ii}\rangle}{\langle L_{ii}\rangle}-1,\quad \beta_{i}:=\frac{\langle L_{ii}(\ell_i-\log  L_{ii})^2\rangle}{\langle L_{ii}\rangle}.\label{eq:sigmoid_para}\end{align}
The parameters in the Gaussian model are justified by the relation between hitting and recurrence times  \cite{HaydnLacroixVaienti2005} on  an ergodic  Markov chain with detailed balance, and become asymptotically exact if   distinct word patterns are statistically independent (such as $ \alpha_{13}$, $ \alpha_{24}$, $ \alpha_{31}$, $ \alpha_{34}$ in Fig.~\ref{fig:aff}a). Here, statistical independence justifies additivity of variances, hence the $ \sqrt{n_{ij}}$ factor in \eqref{eq:sigmoidal_modulation}; sums of independent samples of $ \log L_{ij}$ become asymptotically Gaussian, thanks to the central limit theorem. Failing that, the actual ranking of $ \langle \log L_{ij}\rangle$ may deviate from the Gaussian model prediction in \eqref{eq:sigmoidal_modulation}, such as the intimately related pairs of words \textit{Elizabeth/Darcy}, \textit{Elizabeth/Jane}, \textit{Darcy/Elizabeth}, \textit{Darcy/pride} and \textit{pride/Darcy}.

\subsubsection{Markov criteria for semantic cliques}
Empirically, we find  that  higher  $ \alpha_{ij}(\ell)$   scores point to closer  affinities between word patterns (Fig.~\ref{fig:aff}a), attributable to kinship (\textit{Elizabeth}, \textit{Jane}), courtship (\textit{Darcy}, \textit{Elizabeth}), disposition (\textit{Darcy}, \textit{pride}) and so on. Our robot reader automatically detects such affinities,   without    references other than the novel itself. Therefore, we can use the  $ \alpha_{ij}(\ell)$ scores as guides to  numerical approximations of  semantic fields, hereafter referred to as \textit{semantic cliques}.

We invite a topical pattern  $ \mathsf W_j$ to the semantic clique $ \mathscr S_i$ (Figs.~\ref{fig:aff}a and b, insets) surrounding
$ \mathsf W_i$, if $ \min\{\alpha_{ij}(\langle \log L_{ij}\rangle),\alpha_{ji}(\langle \log L_{ji}\rangle)\}>\alpha_*$ for a standard Gaussian threshold  $\alpha_*:=\frac{1}{\sqrt{2\pi}}\int_{-\infty}^1e^{-x^2/2}\mathrm{d}\, x\approx0.8413
$. This operation emulates the brainstorming procedure of a human reader, who associates one word  with another only when they stay much closer than two randomly picked words, according to his/her impression.
\begin{figure*}
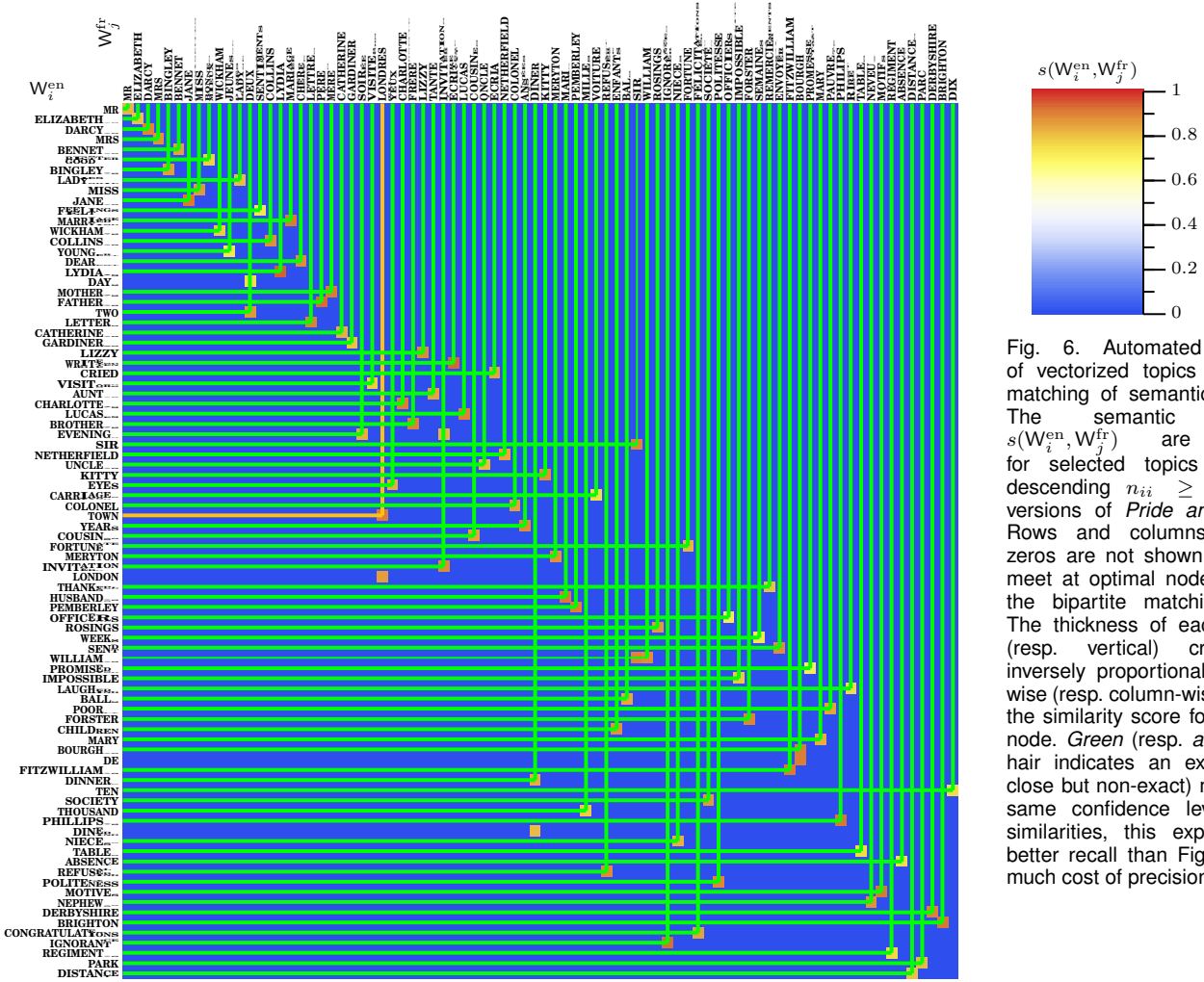

\vspace{-5em}

\begin{minipage}{.76\textwidth}\fontfamily{pnc}\selectfont\begin{center}
\begin{spacing}{0}


\caption{Automated alignments of vectorized topics via bipartite matching of semantic similarities. The semantic similarities  $ s(\mathsf W_i^{\mathrm{en}},\mathsf W_j^{\mathrm{fr}})$ are computed for selected topics (sorted by descending  $ n_{ii}\geq20$) in two versions  of \textit{Pride and Prejudice}.     Rows and columns filled with zeros are not shown. Cross-hairs meet  at optimal nodes  that  solve the bipartite matching problem.  The thickness of each horizontal (resp.~vertical) cross-hair is inversely proportional to the row-wise (resp.~column-wise) ranking of the similarity score for the optimal node.  \textit{Green} (resp.~\textit{amber}) cross-hair indicates an exact (resp.~a close but non-exact) match.  At the same confidence level (0.7) for similarities, this experiment has better  recall than Fig.~\ref{fig:topic_ext}c, without much cost of precision.
\label{fig:vec_align}}
\end{minipage}

\vspace{4em}
\end{figure*}

 Indeed, by numerical brainstorming from $ \mathsf W_i$, our  semantic cliques $  \mathscr S_i$ (Figs.~\ref{fig:aff}a and b, insets)  inform us about their center word $ \mathsf W_i$, through several types of semantic relations, including, but  not limited to \begin{itemize}
\item
Synonyms (\textit{pride} and \textit{vanity} in English, \textit{orgeuil} and \textit{fiert\'e} in French, etc.);\item Temperaments (\textit{Elizabeth}, a  \textit{delightful} girl, often \textit{laughs}, corresponding to French verbs \textit{sourire} and \textit{rire});\item Co-references (\textlat{e.g.}~\textit{Darcy} as a personification of \textit{pride});\item Causalities (such as \textit{pride} based on \textit{fortune}).
\end{itemize}

On a local graph with vertices  $ \mathscr S_i=\{\mathsf W_{i_1}=\mathsf W_i,$ $\mathsf W_{i_2},$ $\dots,$ $\mathsf W_{i_{N_i}}\}$,   we specify the connectivity of each directed edge by a localized  Markov  matrix $ \mathbf{ P}^{[i]}=( p_{jk}^{[i]})_{1\leq j,k\leq N_i}$. This localized Markov matrix
 is the  row-wise normalization of  an $  N_{i}\times N_{i}$ subblock of  $ \mathbf{ P}$ with the same set of vertices as $ \mathscr S_i$.  Resetting the entries $  p_{1k}^{[i]}$ and $  p^{[i]}_{j1}$ as zero, one arrives at the localized recurrence matrix $ \mathbf{ R}^{[i]}$. We call   $ \mathbf{ R}^{[i]}$ a recurrence matrix, because one can use it to  compute  the distribution for
recurrence times to the Markov state
$ \mathsf W_i$ in $ \mathscr S_i$. As we will see soon in the applications below, the eigenvalues of    $ \mathbf{ R}^{[i]}$, when properly arranged, become language-independent semantic fingerprints.

\section{Applications}

\subsection{Automated word translations from bilingual documents}Experimentally, we  resolve the connectivity  of an individual pattern
$ \mathsf W_i$ through  the recurrence spectrum  $ \sigma(\mathbf{ R}^{[i]})$ (Fig.~\ref{fig:aff}b).
The dominant eigenvalues
of $ \mathbf{ R}^{[i]}$ are concept-specific while remaining nearly language-independent (a localized version of the invariance in Fig.~\ref{fig:Markov_model}c).
Such empirical evidence motivates us to define the language-independent \textit{semantic fingerprint} (LISF) of a word  pattern $\mathsf W_i$
 by a descending list for the magnitudes of  eigenvalues \begin{align}\mathbf v_i=(|\lambda_1( \mathbf{ R}^{[i]})|,|\lambda_2( \mathbf{ R}^{[i]})| , \dots),\label{eq:LISF}\end{align}computable from its semantic clique $ \mathscr S_i$. We zero-pad this vector from the   $ (\lfloor e^{\eta_i}\rfloor+1)$st component onwards, where
 $ \eta_i$ is the Kolmogorov--Sinai entropy production rate of the Markov matrix $\mathbf{ P}^{[i]}$, measured in nats per word.\footnote{The entropy production rate   $ \eta(\mathbf P):=-\sum_{i,j}\pi_ip_{ij}\log p_{ij}$   \cite[(4.27)]{CoverThomas} of a Markov matrix $ \mathbf P$ represents the weighted average (assigning probability mass  $\pi_i$ to the \textit{i}th Markov state) of Boltzmann's partition entropies $ -\sum_{j}p_{ij}\log p_{ij}$    \cite[\S8.2]{PollicottYuri}. We have $\eta(\mathbf P)\leq \log N $  for an $ N\times N$  Markov matrix $ \mathbf P$ with strictly positive entries\cite[Theorem~14.1]{PollicottYuri}.}

Via  bipartite matching  (Fig.~\ref{fig:vec_align}) of word vectors $ \mathbf v_i$ across languages, our algorithm translates words   from parallel texts at very high precision  (Fig.~\ref{fig:aff}c), being competitive with state-of-the-art algorithms for bilingual word mapping \cite{Joulin2018,Chen2018}.

Unlike the vector $\mathbf b_i$ (Fig.~\ref{fig:topic_ext}c) that captures only chapter-scale features of $\mathsf W_i$, the semantic fingerprint  $ \mathbf v_i$ in \eqref{eq:LISF} characterizes the kinetic  behavior of $\mathsf W_i$ on all the long-range time scales.

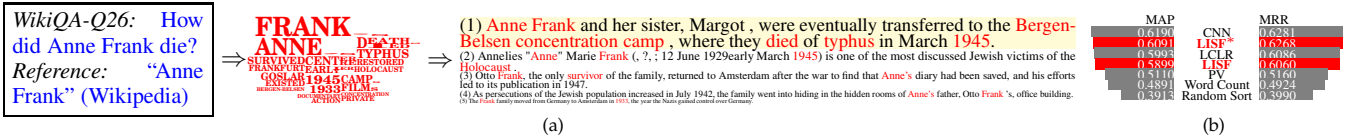
\begin{figure*}

\hspace{1em}\fbox{\begin{minipage}{.14\textwidth}
 \fontsize{0.3cm}{1em}\fontfamily{ptm}\selectfont \textit{WikiQA-Q26: }
\textcolor{blue}{How did Anne Frank die?}\\ \textit{Reference:}
\textcolor{blue}{  ``Anne Frank'' (Wikipedia)}\end{minipage}}\hspace{.3em}$\Rightarrow $\hspace{.08em}
\begin{minipage}{.102\textwidth}\fontfamily{pnc}\selectfont\begin{spacing}{0}
\begin{picture}(98.998,53.04)(-25,-40){\textbf{\put(-25.411,-4.246){\textcolor{red}{
\setlength{\unitlength}{0.255cm}{
\BC{0}{0.}{5}{1.}{FRANK}
\BC{5}{0.}{1}{0.099}{'}
\BC{6}{0.}{1}{0.099}{S}
}}}
\put(-25.411,-11.722){\textcolor{red}{
\setlength{\unitlength}{0.224cm}{
\BC{0}{0.}{4}{1.}{ANNE}
\BC{4}{0.}{1}{0.119}{'}
\BC{5}{0.}{1}{0.119}{S}
}}}
\put(12.944,-8.995){\textcolor{red}{
\setlength{\unitlength}{0.142cm}{
\BC{0}{0.}{1}{1.}{D}
\BC{1}{0.}{2}{0.423}{I}
\BC{1}{0.423}{2}{0.577}{EA}
\BC{3}{0.}{1}{0.423}{E}
\BC{3}{0.423}{1}{0.577}{T}
\BC{4}{0.}{1}{0.346}{D}
\BC{4}{0.423}{1}{0.577}{H}
\BC{5}{0.423}{1}{0.115}{S}
}}}
\put(12.944,-12.857){\textcolor{red}{
\setlength{\unitlength}{0.116cm}{
\BC{0}{0.}{6}{1.}{TYPHUS}
}}}
\put(-6.535,-15.52){\textcolor{red}{
\setlength{\unitlength}{0.114cm}{
\BC{0}{0.}{4}{1.}{CENT}
\BC{4}{0.}{1}{0.5}{R}
\BC{4}{0.5}{1}{0.5}{E}
\BC{5}{0.}{1}{0.5}{E}
\BC{5}{0.5}{1}{0.5}{R}
}}}
\put(-28.324,-14.908){\textcolor{red}{
\setlength{\unitlength}{0.096cm}{
\BC{0}{0.}{8}{1.}{SURVIVED}
}}}
\put(-27.887,-17.682){\textcolor{red}{
\setlength{\unitlength}{0.083cm}{
\BC{0}{0.}{9}{1.}{FRANKFURT}
}}}
\put(-6.535,-18.608){\textcolor{red}{
\setlength{\unitlength}{0.093cm}{
\BC{0}{0.}{4}{1.}{EARL}
\BC{4}{0.}{1}{0.545}{Y}
\BC{4}{0.545}{1}{0.455}{I}
\BC{5}{0.545}{1}{0.455}{E}
\BC{6}{0.545}{1}{0.455}{R}
}}}
\put(-23.28,-20.947){\textcolor{red}{
\setlength{\unitlength}{0.098cm}{
\BC{0}{0.}{6}{1.}{GOSLAR}
}}}
\put(-6.535,-22.568){\textcolor{red}{
\setlength{\unitlength}{0.119cm}{
\BC{0}{0.}{4}{1.}{1945}
}}}
\put(7.005,-22.104){\textcolor{red}{
\setlength{\unitlength}{0.105cm}{
\BC{0}{0.}{4}{1.}{CAMP}
\BC{4}{0.}{1}{0.214}{S}
}}}
\put(12.944,-15.496){\textcolor{red}{
\setlength{\unitlength}{0.079cm}{
\BC{0}{0.}{8}{1.}{RESTORED}
}}}
\put(7.005,-25.41){\textcolor{red}{
\setlength{\unitlength}{0.099cm}{
\BC{0}{0.}{4}{1.}{FILM}
\BC{4}{0.}{1}{0.5}{S}
}}}
\put(11.952,-17.943){\textcolor{red}{
\setlength{\unitlength}{0.073cm}{
\BC{0}{0.}{9}{1.}{HOLOCAUST}
}}}
\put(-4.917,-26.054){\textcolor{red}{
\setlength{\unitlength}{0.105cm}{
\BC{0}{0.}{4}{1.}{1933}
}}}
\put(-21.503,-23.448){\textcolor{red}{
\setlength{\unitlength}{0.075cm}{
\BC{0}{0.}{7}{1.}{EXISTED}
}}}
\put(-24.82,-25.093){\textcolor{red}{
\setlength{\unitlength}{0.049cm}{
\BC{0}{0.}{13}{1.}{BERGEN-BELSEN}
}}}
\put(7.005,-27.032){\textcolor{red}{
\setlength{\unitlength}{0.049cm}{
\BC{0}{0.}{13}{1.}{CONCENTRATION}
}}}
\put(-9.461,-27.804){\textcolor{red}{
\setlength{\unitlength}{0.052cm}{
\BC{0}{0.}{11}{1.}{DOCUMENTARY}
}}}
\put(7.005,-29.169){\textcolor{red}{
\setlength{\unitlength}{0.064cm}{
\BC{0}{0.}{7}{1.}{PRIVATE}
}}}
\put(-4.115,-29.972){\textcolor{red}{
\setlength{\unitlength}{0.065cm}{
\BC{0}{0.}{6}{1.}{ACTION}
}}}
}}\end{picture}\end{spacing}\end{minipage}\hspace{1.23em}$\Rightarrow $\hspace{.25em}\begin{minipage}{0.45\textwidth}\begin{picture}(1,1)(0,32)\textcolor{yellow!20!white}{\polygon*(0,20)(233,20)(233,32)(0,32)}\end{picture}
\fontfamily{ptm}\selectfont{\begin{spacing}{0}
{\fontsize{0.25112cm}{1em}\selectfont (1) \textcolor{red}{Anne} \textcolor{red}{Frank} and her sister, Margot , were eventually transferred to the \textcolor{red}{Bergen-Belsen} \textcolor{red}{concentration} \textcolor{red}{camp} , where they \textcolor{red}{died} of \textcolor{red}{typhus} in March \textcolor{red}{1945}. }

{\fontsize{0.16709cm}{1em}\selectfont (2) Annelies "\textcolor{red}{Anne}" Marie \textcolor{red}{Frank} (, ?, ; 12 June 1929early March \textcolor{red}{1945}) is one of the most discussed Jewish victims of the \textcolor{red}{Holocaust} . }

{\fontsize{0.14577cm}{1em}\selectfont (3) Otto \textcolor{red}{Frank}, the only \textcolor{red}{survivor} of the family, returned to Amsterdam after the war to find that \textcolor{red}{Anne's} diary had been saved, and his efforts led to its publication in 1947. }

{\fontsize{0.12496cm}{1em}\selectfont (4) As persecutions of the Jewish population increased in July 1942, the family went into hiding in the hidden rooms of \textcolor{red}{Anne's} father, Otto \textcolor{red}{Frank} 's, office building. }

{\fontsize{0.08378cm}{1em}\selectfont (5) The \textcolor{red}{Frank} family moved from Germany to Amsterdam in \textcolor{red}{1933}, the year the Nazis gained control over Germany. }
\end{spacing}}\end{minipage}\begin{minipage}{.1\textwidth}{
\renewcommand{\arraystretch}{0.5}\begin{center}
\vspace{-.1em}
\begin{tiny}
\hspace{-4.5em}\begin{tabular}{r@{\;\;}c@{\;\;}l}
{\tiny \fontfamily{ptm}\selectfont MAP}&{\tiny\textit{}}&{\tiny \fontfamily{ptm}\selectfont MRR\textit{}}\\
\textcolor{gray}{\rule{30.95pt}{4pt}}\hspace{-13.5pt}\textcolor{white}{\tiny\fontfamily{ptm}\selectfont0.6190}&{\tiny \fontfamily{ptm}\selectfont CNN}&\textcolor{gray}{\rule{31.41pt}{4pt}}\hspace{-31.41pt}\textcolor{white}{\tiny\fontfamily{ptm}\selectfont0.6281}\\[-1pt]
\textcolor{red}{\rule{30.43pt}{4pt}}\hspace{-13.5pt}\textcolor{white}{\tiny\fontfamily{ptm}\selectfont0.6091}&\textbf{\tiny\textcolor{red}{\fontfamily{ptm}\selectfont LISF$^*$}}&\textcolor{red}{\rule{31.31pt}{4pt}}\hspace{-31.31pt}\textcolor{white}{\tiny\fontfamily{ptm}\selectfont0.6268}\\[-1pt]
\textcolor{gray}{\rule{29.97pt}{4pt}}\hspace{-13.5pt}\textcolor{white}{\tiny\fontfamily{ptm}\selectfont0.5993}&{\tiny \fontfamily{ptm}\selectfont LCLR}&\textcolor{gray}{\rule{30.43pt}{4pt}}\hspace{-30.43pt}\textcolor{white}{\tiny\fontfamily{ptm}\selectfont0.6086}\\[-1pt]
\textcolor{red}{\rule{29.49pt}{4pt}}\hspace{-13.5pt}\textcolor{white}{\tiny\fontfamily{ptm}\selectfont0.5899}&\textbf{\tiny\textcolor{red}{\fontfamily{ptm}\selectfont LISF}}&\textcolor{red}{\rule{30.3pt}{4pt}}\hspace{-30.3pt}\textcolor{white}{\tiny\fontfamily{ptm}\selectfont0.6060}\\[-1pt]
\textcolor{gray}{\rule{25.55pt}{4pt}}\hspace{-13.5pt}\textcolor{white}{\tiny\fontfamily{ptm}\selectfont0.5110}&{\tiny \fontfamily{ptm}\selectfont PV}&\textcolor{gray}{\rule{25.80pt}{4pt}}\hspace{-25.80pt}\textcolor{white}{\tiny\fontfamily{ptm}\selectfont0.5160}\\[-1pt]
\textcolor{gray}{\rule{24.46pt}{4pt}}\hspace{-13.5pt}\textcolor{white}{\tiny\fontfamily{ptm}\selectfont0.4891}&{\tiny \fontfamily{ptm}\selectfont Word Count}&\textcolor{gray}{\rule{24.62pt}{4pt}}\hspace{-24.62pt}\textcolor{white}{\tiny\fontfamily{ptm}\selectfont0.4924}\\[-1pt]
\textcolor{gray}{\rule{19.57pt}{4pt}}\hspace{-13.5pt}\textcolor{white}{\tiny\fontfamily{ptm}\selectfont0.3913}&{\tiny \fontfamily{ptm}\selectfont Random Sort}&\textcolor{gray}{\rule{19.95pt}{4pt}}\hspace{-19.95pt}\textcolor{white}{\tiny\fontfamily{ptm}\selectfont0.3990}
\end{tabular}\end{tiny}\end{center}}\end{minipage}\begin{minipage}{.78\textwidth}

\end{minipage}

\vspace{-.5em}
\begin{minipage}{.84\textwidth}\begin{center}{\scriptsize(a)}\end{center}\end{minipage}\begin{minipage}{.125\textwidth}\begin{center}{\scriptsize(b)}\end{center}\end{minipage}

\vspace{-1em}
\caption{Applications of semantic cliques  to question-answering. (a)~A construction of semantic clique  $ \mathscr Q\cup\mathscr Q'$ (based on $\mathscr Q=\{ $\textit{Anne}, \textit{Frank}, \textit{die}$ \}$)  weighted by the PageRank equilibrium state $ \bm {\widetilde \pi}$  and subsequent question-answering. Top 5 candidate answers, with punctuation and spacing as given by WikiQA, are shown with font sizes proportional to  the  entropy production  score  in \eqref{eq:sentence_score}. Here, the top-scoring sentence with \textit{highlighted} background is the same as  the official answer chosen by  the WikiQA team. Like a human reader, our algorithm automatically detects the  place ``Bergen-Belsen concentration camp'', cause ``typhus'', and year ``1945'' of Anne Frank's death. (b)~Evaluations of our model (LISF and LISF$^*$) on the WikiQA data set, in comparison with  established algorithms.\label{fig:QA}}\vspace{-2em}
\end{figure*}

Given a topical pattern $ \mathsf W_i^{\mathrm A}$ in language A, its semantic fingerprint $ \mathbf {v}_i^{\mathrm A}$ (a descending list of recurrence eigenvalues, as in Fig.~\ref{fig:aff}b) allows us to  numerically locate a semantically close pattern in a parallel text   written in  another language B, in two steps: \\ (1)~Divide the document  into $ K$ chapters, and define the semantic similarity function as  $ s(\mathsf W_i^{\mathrm A},\mathsf W_j^{\mathrm B}):=s_R({\mathbf v}_i^{\mathrm A},{\mathbf v}_j^{\mathrm B})$
  if \begin{align}  s_R(\mathbf b_i^{\mathrm A},\mathbf b_j^{\mathrm B})\geq\max\left\{1-0.07\sqrt{K},1-\sqrt{\frac{\Vert \smash[b]{\mathbf b_i^{\mathrm A}\wedge\mathbf b_j^{\mathrm B}}\Vert_0}{\Vert \smash[b]{\mathbf b_i^{\mathrm A}\vee\mathbf b_j^{\mathrm B}}\Vert_1}}\right\}\end{align} (which is a ballpark screening  more robust  than Fig.~\ref{fig:topic_ext}c, with  $ \Vert\mathbf b\Vert_0$ counting the number of non-zero components in $ \mathbf b$) and $ s_R({\mathbf v}_i^{\mathrm A},{\mathbf v}_j^{\mathrm B})\geq0.7$;  $ s(\mathsf W_i^{\mathrm A},\mathsf W_j^{\mathrm B}):=0$ otherwise.  \\ (2)~Solve a bipartite matching problem (Fig.~\ref{fig:vec_align}) that maximizes  $\sum_{i,j} s(\mathsf W_i^{\mathrm A},\mathsf W_j^{\mathrm B})$, using the Hungarian Method \cite{Kuhn1955} attributed to Jacobi--K\H{o}nig--Egerv\'ary--Kuhn \cite{Kuhn2012}.

\subsection{Machine-assisted text comprehension on WikiQA data set}

 By automatically discovering related words through numerical brainstorming (Figs.~\ref{fig:aff}a and b, insets),  our semantic cliques  $  \mathscr S_i$ are useful  in text comprehension and question answering. We can expand a set of question words $ \mathscr Q=\{\mathsf W_{q_1},\dots,\mathsf W_{q_K}\}$ into    $ \mathscr Q\cup\mathscr Q'=\bigcup_{k=1}^K\mathscr S_{q_k}$, by bringing together the semantic cliques $ \mathscr S_{q_k}$ generated from a reference text by each and every   question word $ \mathsf W_{q_k}$.

As before, we  construct a localized Markov matrix   $ \mathbf { P}=(p_{ij})_{1\leq i,j\leq N}$ on this subset of word patterns  $ \mathscr Q\cup\mathscr Q'$. We further use the Brin--Page damping \cite{BrinPage1998} to derive an ergodic Markov matrix $ \mathbf {\widetilde  P}=(\widetilde p_{ij})_{1\leq i,j\leq N}$, where $ \widetilde p_{ij}=0.85p_{ij}+\frac{0.15}{N}$.

By analogy to the behavior of internet surfing \cite{BrinPage1998,PageRank}, we model the process of associative reasoning \cite{Sloman1996} as a navigation through the nodes $\mathscr Q\cup\mathscr Q' $ according to  $ \mathbf {\widetilde P}$, which quantifies the click-through rate from one idea to another.
The Page\-Rank recursion \cite{PageRank} ensures a unique equilibrium state  $\bm{ \widetilde \pi}$ attached to $ \mathbf {\widetilde P}$. If our question $Q$ and a candidate answer  $A$ contain, respectively,  words from    $ \mathsf W_{Q_1}$, $\dots$, $\mathsf W_{Q_{m}}\in  \mathscr Q$ and  $ \mathsf W_{A_1}$, $\dots$, $\mathsf W_{A_{n}}\in  \mathscr Q\cup\mathscr Q' $ (counting multiplicities, but excluding function words and patterns with fewer than 3 occurrences in the reference document), then we assign the following entropy production score \begin{align}\mathscr F[Q,A]:=-\sum_{i=1}^m
\sum_{j=1}^{n}\widetilde \pi_{Q_i}\widetilde p_{Q_iA_j}\log\widetilde p_{Q_iA_j}\label{eq:sentence_score}\end{align}to  this question-answer pair.\footnote{One may compare the score $ \mathscr F[Q,A]$  to the Kolmogorov--Sinai entropy production rate \cite[(4.27)]{CoverThomas} $ \eta(\mathbf P)= -\sum_{i=1}^N\sum_{j=1}^N\pi_i p_{ij}\log p_{ij}$ of a Markov matrix $ \mathbf P=(p_{ij})_{1\leq i,j\leq N}$. The score  $ \mathscr F[Q,A] $  is modeled after Boltzmann's partition entropies, weighted by words in the question, and sifted by topics in the answer. Such a weighting and sifting method is analogous to the definition of scattering cross-sections in particle physics.}

 A sample work flow  is shown in Fig.~\ref{fig:QA}a,
to illustrate how our rudimentary question-answering machine  handles a query. To answer a question, we use    a single Wikipedia page (without infoboxes and other structural data)  as the only reference document and  training source.  Like a typical human reader of Wikipedia, our numerical associative reasoning generates a weighted set of nodes  $ \mathscr Q\cup\mathscr Q'$ (presented graphically as a thought bubble in Fig.~\ref{fig:QA}a), without the help of  external stimuli or knowledge feed. Here, the relative weights   in the nodes of   $ \mathscr Q\cup\mathscr Q'$  are computed from  the  equilibrium state  $\bm{ \widetilde \pi}$ of $ \mathbf {\widetilde P}$, via the  PageRank algorithm.

 We then  test our semantic model  (LISF in Fig.~\ref{fig:QA}b) on all the 1242 questions   in the   Wiki\-QA data set, each of which is accompanied by at least one correct
answer located in a designated Wikipedia page.
Our  algorithm's performance is  roughly on par with  LCLR   and CNN   benchmarks \cite{WikiQA},   improving upon the baseline by significant margin. This is perhaps remarkable, considering the relatively scant data at our disposal. Unlike the LCLR approach, our numerical discovery of synonyms does not draw on the WordNet database \cite{WordNet} or pre-existent corpora of question-answer pairs. Unlike the CNN\ method, we do not need pre-trained word2vec embeddings \cite{NIPS2013word2vec} as semantic input.

Moreover, our algorithm  (LISF$^* $ in Fig.~\ref{fig:QA}b) performs slightly better on a subset of 990 questions that do not require quantitative cues (\textit{How big? How long? How many? How old? What became of? What happened to? What year?} and so on). This indicates that, with a Markov chain description of two-body interactions between topics,  our structural model fits associative reasoning better than rule-based reasoning \cite{Sloman1996}, while imitating human behavior in the presence of limited data. To enhance the reasoning capabilities of our algorithm, it is perhaps appropriate to apply a Markov random field \cite[\S4.1.3]{Mumford2010} to graphs of word patterns, to capture many-body interactions among different topics.

\section{Conclusion}
In our current work, we  define semantics through algebraic invariants that are concept-specific and language-independent. To construct such invariants, we develop a stochastic model 
that assigns a semantic fingerprint (list of recurrence eigenvalues) to each
concept via its long-range contexts. Consistently using a  single  Markov framework,  we are able  to     extract topics (Figs.~\ref{fig:rec_topic}e, \ref{fig:topic_ext}b,b'), translate topics (Figs.~\ref{fig:topic_ext}c, \ref{fig:Markov_model}c, \ref{fig:aff}b,c, \ref{fig:vec_align}) and understand topics (Figs.~\ref{fig:aff}a,b,  \ref{fig:QA}a,b), through statistical mining of short and medium-length texts.  In view of these three successful applications, we are probably close to a complete set of semantic invariants, after demystifying the long-range behavior of human languages.

Notably, our algorithms apply to documents of moderate lengths, similar to  the experience of  human readers. This contrasts with data-hungry algorithms in machine learning \cite{WikiQA,Tshitoyan2019}, which utilize high-dimensional numerical representations of words and phrases \cite{NIPS2013word2vec,Arora2016,Joulin2018,Chen2018} from large corpora. Our  semantic mechanism exhibits universality on long-range linguistic scales. This adds to our quantitative understanding of diversity on shorter-range linguistic scales, such as phonology \cite{NowakKrakauer1999,EBR2015,Everett2017}, morphology \cite{Pinker1997WordsRules,MarslenWilson1997,Lieberman2007,Newberry2017} and  syntax \cite{PinkerSurv2000,NowakPlotkinJansen2000,Chomsky2002syntactic,DunnGreenhillLevinsonGray2011,Newberry2017}.

Thanks to the independence between semantics and syntax \cite{Chomsky2002syntactic}, our current model conveniently ignores the non-Markovian syntactic structures which are essential to  fluent speech.
In the near future, we
hope to extend our framework further, to incorporate both Markovian and non-Markovian features across different ranges. \textit{The Mathematical Principles of Natural Languages}, as we envision, must and will combine the statistical analysis  of a Markov model   with linguistic properties on shorter time scales that convey morphological \cite{Pinker1997WordsRules,MarslenWilson1997,Lieberman2007,Newberry2017} and syntactical \cite{PinkerSurv2000,NowakPlotkinJansen2000,Chomsky2002syntactic,DunnGreenhillLevinsonGray2011,Newberry2017} information.


%

\appendices
\section{ Poissonian banality and non-topicality\label{app:two-sigma_Poisson}}
Here, we first present a proof of our statistical criterion for Poissonian banality (which we identify with non-topicality of word patterns), as stated in \eqref{ineq:95_conf_large_N}.
\begin{theorem}[Sums and products of exponentially distributed random variables]\label{thm:YN_dist}Let $ X_1,\dots,X_N$ be independent random variables, each obeying an exponential distribution with mean 1.  The probability distribution of
\begin{align} Y_N:=\log\left(\frac{1}{N}\sum_{i=1}^NX_i\right)-\frac{1}{N}\sum_{i=1}^N\log X_i\end{align}is asymptotic to a Gaussian law\begin{align}\mathbb P\left( \frac{Y_N-\left( \gamma_0-\frac{1}{2N} \right)}{\frac{1}{\sqrt{N}}\sqrt{\frac{\pi^2}{6}-1-\frac{1}{2N}}}<a\right)\sim\int_{-\infty}^ae^{-{x^2}/{2}}\frac{\D x}{\sqrt{2\pi}}\label{eq:central_limit}
\end{align}as $ N\to\infty$.\end{theorem}\begin{IEEEproof}By definition, we have a moment-generating function \begin{align}&
\mathbb E e^{-t Y_N}\notag\\={}&\int_{(0,+\infty)^N}\frac{\prod_{i=1}^Nx_i^{t/N}}{\left(\frac{1}{N}\sum_{i=1}^Nx_i\right)^t}e^{-\sum_{i=1}^Nx_i}\D x_1\cdots \D x_N\notag\\={}&2^N\int_{(0,+\infty)^N}\frac{\prod_{i=1}^N\xi_i^{2t/N+1}}{\left(\frac{1}{N}\sum^N_{i=1}\xi_{i}^{2}\right)^t}e^{-\sum^N_{i=1}\xi_{i}^{2}}\D \xi_{1}\cdots\D \xi_{N}.
\end{align}To evaluate the last multiple integral, we use  the spherical coordinates $ \xi_1=r\cos\theta_1$, $ \xi_2=r\sin\theta_1\cos\theta_2$, $ \xi_3=r\sin\theta _1\sin\theta_2\cos\theta_3$, $ \dots$, $ \xi_N=r\prod_{j=1}^{N-1}\sin \theta_j$ (where $ r>0$, and $ 0<\theta_j<\pi/2$ for all $ j\in\{1,\dots, N-1\}$), with volume element \begin{align} \D \xi_{1}\cdots\D \xi_{N}=r^{N-1}\D r\prod_{j=1}^{N-1}\sin^{N-1-j} \theta_j\D\theta_j.\end{align} The result reads\begin{align}
\mathbb E e^{-t Y_N}={}&N^{t}\Gamma(N)\prod_{j=1}^{N-1}\frac{\Gamma\left(\frac{N+t}{N}\right)\Gamma\left(\frac{(N-j)(N+t)}{N}\right)}{\Gamma\left(\frac{(N-j+1)(N+t)}{N}\right)}\notag \\={}& \frac{N^{t}\Gamma(N)}{\Gamma(N+t)}\left[ \Gamma\left( 1+\frac{t}{N} \right) \right]^N,\label{eq:Y_N_Lap}\end{align}where $ \Gamma(s):=\int_0^\infty x^{s-1}e^{-x}\D x$ is Euler's gamma function.


Consequently, the proof of \eqref{eq:central_limit} builds on a cumulant expansion of  \eqref{eq:Y_N_Lap}, that is, development of $ \log
\mathbb E e^{-t Y_N}$ up to $ O(t^2)$ terms.\end{IEEEproof}

If we have $N$ samples of recurrence times from a Poisson process, then the statistic $Y_N$ satisfies the inequality in  \eqref{ineq:95_conf_large_N}  with probability $ \int_{-2}^2e^{-{x^2}/{2}}\frac{\D x}{\sqrt{2\pi}}\approx0.95$, in view of the theorem above.


\section{Recurrence times on an ergodic Markov chain with detailed balance}

\subsection{Background in probability theory}
Based on numerical evidence (Fig.~\ref{fig:Markov_model}),  we postulate that at the \textit{discourse level} (the  longest time scale in Friederici's   \cite{Friederici1999} neurobiological hierarchy), the production of natural  language texts   can be caricatured by the stochastic transitions on a stationary and ergodic Markov chain $ \mathscr M=(\mathscr S,\mathbf P)$.\footnote{To reduce notational burden, we will not use superscripted asterisks to mark  Markov  matrices,  beyond this point.
}
Here, the state space $ \mathscr S=\{\mathsf W_1,\dots,\mathsf W_N\}$ runs over finitely many word patterns occurring in the text, which in turn is representable as a discrete-time stochastic process $ (X(0)$, $X(1)$, $\dots$, $X(n)$, $\dots)$;  the transition matrix $ \mathbf P=(p_{ij})$ describes the conditional hopping  probability on the web of
word patterns: $ p_{ij}=\mathbb P(X(n+1)=\mathsf W_j|X(n)=\mathsf W_i)$, for $ n\in\mathbb Z_{\geq0}$. Depending on context, we also model a document by a localized  Markov chain  $ \mathscr M'=(\mathscr S',\mathbf P')$, where certain word patterns $ \mathsf W_i$ are removed from the state space $ \mathscr S$ to form a proper subset $ \mathscr S'\subsetneqq\mathscr S$.

In a more formal setting, our notation  $ \mathscr  M=(\mathscr S,\mathbf P) $ for the Markov chain should be expanded into $ \mathscr M=(\varOmega,\mathscr F, \{\mathbb P^{\mathsf W}:\mathsf W\in\mathscr S\},\{X(t):t\in\mathbb Z_{\geq0}\}, \{\widehat{\theta}_n:n\in\mathbb  Z_{\geq0}\})$, whose components  are explained below.\begin{itemize}[\IEEEsetlabelwidth{Z}]
\item
The sample space $ \varOmega$ consists of all stochastic trajectories $ \omega=(X(0),X(1),\dots,X(t),X(t+1),\dots)=(X(t))_{t\in\mathbb Z_{\geq0}}$ on the state space $ \mathscr S$, namely,    all possible texts that can be analyzed by our particular model.\item Each member  in the field of events  $ \mathscr F=2^{\varOmega}$ (the totality of all subsets in the sample space  $ \varOmega$) is a set containing zero or more stochastic processes that can be regarded as  caricatures of  text productions. \item The family of probability measures $\{\mathbb P^{\mathsf W}:\mathsf W\in\mathscr S\} $ are related to the transition matrix $ \mathbf P=(p_{ij})_{1\leq i, j\leq N}$ by the identity $ p_{ij}^{(t)}=\mathbb P^{\mathsf W_i}(X(t)=\mathsf W_j)=\mathbb P(X(t)=\mathsf W_j|X(0)=\mathsf W_i),\forall \mathsf W_i,\mathsf W_j\in\mathscr S,\forall t\in\mathbb Z_{>0}$. \item The shift operator $ \widehat{\theta}_n:\varOmega\longrightarrow\varOmega$ acts on an arbitrary trajectory $ (X(0),X(1),\dots,X(t),X(t+1),\dots)=\omega\in\varOmega$ in the following manner: $ \widehat{\theta}_n\circ\omega=(X(n),X(n+1),\dots,X(t+n),X(t+n+1),\dots),\forall t,n\in\mathbb Z_{\geq0}$.
\end{itemize}

 For discussions of the stopping times (a class of random variables on Markov chains) as well as  the  Markov property, it is convenient to further introduce the notation $ \mathscr F_n$ for $ n\in\mathbb Z_{\geq0}$. Here, $\mathscr F_n $ is the smallest $ \sigma$-algebra containing all the events in the form of  $ (X(0)=\mathsf W_{i_0},X(1)=\mathsf W_{i_1},\dots,X(n)=\mathsf W_{i_n})$. The family of $ \sigma$-algebras $ \{\mathscr F_n:n\in\mathbb Z_{\geq0}\}$  forms a filtration: $ \mathscr F_n\subseteq\mathscr F_m$ if $ n\leq m$. For any $ n\in\mathbb Z_{\geq0}$,  $B\in\mathscr F$, $ \mathsf W\in\mathscr S$, we have the following relation concerning conditional expectations: \begin{align} \mathbb E^{\mathsf W}(\mathbf1_{B}\circ\widehat \theta_n|\mathscr F_n)=\mathbb E^{X(n)}(\mathbf1_{B}):=\mathbb P^{X(n)}(B),\label{eq:weak_Markov}\end{align} (for every time-step $ n\in\mathbb Z_{\geq0}$ and Markov state $ \mathsf W\in\mathscr S$) which is merely a reformulation of the  Markov condition $ \mathbb P^{\mathsf W}(X(n+1)=\mathsf W_{i_{n+1}}|X(n)=\mathsf W_{i_n},X(n-1)=\mathsf W_{i_{n-1}},\dots,X(0)=\mathsf W_{i_{0}})=\mathbb P^{\mathsf W_{i_n}}(X(1)=\mathsf W_{i_{n+1}})=\mathbb P^{\mathsf W}(X(n+1)=\mathsf W_{i_{n+1}}|X(n)=\mathsf W_{i_n})$.
In other words, for any $ A\in\mathscr F_n$,  $ B\in\mathscr F$, $\mathsf W\in\mathscr S$, we have the following statement of the \textit{Markov property}:\begin{align}\mathbb E^{\mathsf W}(\mathbf1_B\circ\widehat\theta_n;A)=\mathbb E^{\mathsf W}(\mathbb E^{X(n)}(\mathbf1_B);A).
\label{eq:weak_Markov'}\tag{\ref{eq:weak_Markov}\ensuremath{'}}
\end{align}In both \eqref{eq:weak_Markov} and \eqref{eq:weak_Markov'}, one can replace the indicator function  $ \mathbf1_B$ for event $B\in\mathscr F$ by any random variable with finite expectation.
\subsection{Hitting time and return time on a Markov chain}We need to first precisely define the probability distributions for the hitting time and return time (the latter also known as ``recurrence time'') on our Markov chain $ \mathscr M=(\mathscr S,\mathbf P)$.

For each state $ \mathsf W_i\in\mathscr S$ and trajectory $\omega=(X(t))_{t\in\mathbb Z_{\geq0}}\in\varOmega $, we define \begin{align} \tau_i(\omega):=\inf\{n\in\mathbb Z_{>0}:X(n)=\mathsf W_i\},\end{align} then  $ \tau_{i}:\varOmega\longrightarrow\mathbb Z_{>0}\cup\{+\infty\}$ is a stopping time. Suppose that our pattern  of interest corresponds to a subset of states $ \mathscr W\subseteq\mathscr S$ on the web of words. We  define another stopping time $ \tau_{\mathscr W}:\varOmega\longrightarrow\mathbb Z_{>0}\cup\{+\infty\}$ as\begin{align}
\tau_{\mathscr W}(\omega):={}&\inf\{n\in\mathbb Z_{>0}:X(n)\in \mathscr W\}\notag\\={}&\inf\{\tau_i(\omega):\mathsf W_i\in\mathscr W\}.
\end{align}Clearly, $ \tau_{\mathscr W}(\omega)$ is equal to  the first time when a forward stepwise search lands on the set of interest $ \mathscr W$.
Recalling the invariant measure  $ \bm\pi=(\pi_1,\dots ,\pi_N)$, we  define the cumulative distribution function (CDF) for the hitting time to the set of patterns  $ \mathscr W$ as\begin{align}
H_{\mathscr W}(t):=\sum_{\mathsf W_i\in\mathscr S}\mathbb P^{\mathsf W_i}(\tau_{\mathscr W}< t)\pi_i,\quad t\in\mathbb Z_{>0}.\label{eq:H_A_defn}
\end{align} Similarly, the CDF for the return time  to the set of patterns  $ \mathscr W$ is defined as \begin{align}
R_{\mathscr W}(t):=\sum_{\mathsf W_i\in\mathscr W}\mathbb P^{\mathsf W_i}(\tau_{  \mathscr W}< t)\frac{\pi_i}{\sum_{\mathsf W_j\in\mathscr W}\pi_j},\quad t\in\mathbb Z_{>0.}\label{eq:R_A_defn}
\end{align}

In the next theorem, we present an identity that connects  hitting and return times of Markov states, which in turn, is a discrete analog of the Haydn--Lacroix--Vaient relation for  continuous-time ergodic dynamical systems  \cite{HaydnLacroixVaienti2005}.

\begin{theorem}[Relation between hitting time and return time distributions]\label{thm:H_R_rln}For a  stationary and ergodic Markov chain  $ \mathscr M=(\mathscr S,\mathbf P)$, and a subset of states $ \mathscr W\subseteq\mathscr S$, we have the following identity regarding the probability distribution of hitting and return times to $ \mathscr W$:\begin{align}H_{\mathscr W}(1)=R_{\mathscr W}(1)=0;\;
\frac{H_{\mathscr W}(t+1)}{\sum_{\mathsf W_j\in\mathscr W}\pi_j}=\sum_{ n=1}^t[1-R_{\mathscr W}(n)]\label{eq:H_R_rln}
\end{align}for all $t\in\mathbb Z_{>0}$.\end{theorem}

\begin{IEEEproof}Clearly, our task is equivalent to the  verification of the following  formula:\begin{align}
\sum_{\mathsf W_i\in\mathscr S}\mathbb P^{\mathsf W_i}(\tau_{\mathscr W}\leq t)\pi_i=\sum_{ n=1}^t\sum_{\mathsf W_j\in\mathscr W}[1-\mathbb P^{\mathsf W_j}(\tau_{  \mathscr W}< t)]\pi_{j}\label{eq:H_R_rln'}\tag{\ref{eq:H_R_rln}$'$}
\end{align}for all $t\in\mathbb Z_{>0}$.

For $ t=1$, the left-hand side of \eqref{eq:H_R_rln'} can be computed as follows:\begin{align}
\sum_{\mathsf W_i\in\mathscr S}\mathbb P^{\mathsf W_i}(\tau_{\mathscr W}=1)\pi_i={}&\sum_{\mathsf W_i\in\mathscr S}\sum_{\mathsf W_j\in\mathscr W}\mathbb P^{\mathsf W_i}(X(1)=\mathsf W_{j})\pi_i\notag\\={}&\sum_{\mathsf W_i\in\mathscr S}\sum_{\mathsf W_j\in\mathscr W}\pi_i p_{ij}=\sum_{\mathsf W_j\in\mathscr W}\pi_j.\label{eq:H_R_rln_1}
\end{align}This is equal to the right-hand side of  \eqref{eq:H_R_rln'}, because  $ \mathbb P^{\mathsf W_j}(\tau_{  \mathscr W}<1)=0$.

For $ t\in\mathbb Z_{>0}$, we can compute{\allowdisplaybreaks\begin{align}&
\sum_{\mathsf W_i\in\mathscr S}\mathbb P^{\mathsf W_i}(\tau_{\mathscr W}\leq t+1)\pi_i\notag\\={}&\sum_{\mathsf W_i\in\mathscr S}\mathbb E^{\mathsf W_i}((\mathbf1_{(\tau_{\mathscr W}\leq t)}\circ\widehat\theta_1)(\mathbf1_{(X(1)\notin\mathscr W)})+\mathbf1_{(X(1)\in\mathscr W)})\pi_i\notag\\={}&\sum_{\mathsf W_i\in\mathscr S}[\mathbb E^{\mathsf W_i}(\mathbb E^{X(1)}(\mathbf1_{(\tau_{\mathscr W}\leq t)};X(1)\notin\mathscr W))\notag\\{}&+\mathbb E^{\mathsf W_i}(\mathbf1_{(X(1)\in\mathscr W)})]\pi_i
\end{align}}from the Markov property \eqref{eq:weak_Markov'}. Here, we have  $ \sum_{\mathsf W_i\in\mathscr S}\mathbb E^{\mathsf W_i}(\mathbf1_{(X(1)\in\mathscr W)})\pi_i=\sum_{\mathsf W_i\in\mathscr S}\mathbb P^{\mathsf W_i}(\tau_{\mathscr W}=1)\pi_i=\sum_{\mathsf W_j\in\mathscr W}\pi_j$ according to \eqref{eq:H_R_rln_1}, and{\allowdisplaybreaks\begin{align}&
\sum_{\mathsf W_i\in\mathscr S}\mathbb E^{\mathsf W_i}(\mathbb E^{X(1)}(\mathbf1_{(\tau_{\mathscr W}\leq t)};X(1)\notin\mathscr W))\pi_i\notag\\={}&\sum_{\mathsf W_i\in\mathscr S}\sum_{\mathsf W_{k}\in\mathscr S\smallsetminus\mathscr W}\mathbb P^{\mathsf W_i}(X(1)=\mathsf W_{k})\mathbb E^{\mathsf W_{k}}(\mathbf1_{(\tau_{\mathscr W}\leq t)})\pi_i\notag\\={}&\sum_{\mathsf W_{k}\in\mathscr S\smallsetminus\mathscr W} \mathbb E^{\mathsf W_{k}}(\mathbf1_{(\tau_{\mathscr W}\leq t)})\pi_k\notag\\={}&\sum_{\mathsf W_i\in\mathscr S}\mathbb P^{\mathsf W_i}(\tau_{\mathscr W}\leq t)\pi_i-\sum_{\mathsf W_j\in\mathscr W}\mathbb P^{\mathsf W_j}(\tau_{  \mathscr W}\leq t)\pi_{j}.
\end{align}}Therefore, the recursion\begin{align}&
\sum_{\mathsf W_i\in\mathscr S}\mathbb P^{\mathsf W_i}(\tau_{\mathscr W}\leq t+1)\pi_i\notag\\={}&\sum_{\mathsf W_i\in\mathscr S}\mathbb P^{\mathsf W_i}(\tau_{\mathscr W}\leq t)\pi_i+\sum_{\mathsf W_j\in\mathscr W}[1-\mathbb P^{\mathsf W_j}(\tau_{  \mathscr W}\leq t)]\pi_{j}\label{eq:H_R_rec}
\end{align}for all $t\in\mathbb Z_{>0}$ allows us to  build \eqref{eq:H_R_rln'}  inductively on the  $ t=1$ case \eqref{eq:H_R_rln_1}.   \end{IEEEproof}
\subsection{Positive definite return time distributions  on a Markov chain with detailed balance\label{app:rec_time}}Thanks to Theorem~\ref{thm:H_R_rln}, our analysis of a return time distribution can be built on the analysis of the corresponding hitting time, the latter of which is usually easier to compute (in theoretical analysis).

For a   stationary and ergodic Markov chain $ (\mathscr S,\mathbf P=(p_{ij})_{1\leq  i, j\leq N})$, and a pattern of interest  $ \mathscr W\subsetneqq\mathscr S$, one can use the Markov property to show that the hitting time distribution satisfies\begin{align}
&\sum_{\mathsf W_{i}\in\mathscr S}\mathbb P^{\mathsf W_i}(\tau_{\mathscr W}>1) \pi_i=\sum_{\mathsf W_{i}\in\mathscr S}\mathbb P^{\mathsf W_i}(X(1)\notin\mathscr W) \pi_i\notag\\={}&\sum_{\mathsf W_{i}\in\mathscr S}\sum_{\mathsf W_{m}\in\mathscr S\smallsetminus\mathscr W}\pi_ip_{i m}=\sum_{\mathsf W_{m}\in\mathscr S\smallsetminus\mathscr W}\pi_{ m}\label{eq:Markov_hitting_1}
\end{align}and {\allowdisplaybreaks\begin{align}&
\sum_{\mathsf W_{i}\in\mathscr S}\mathbb P^{\mathsf W_i}(\tau_{\mathscr W}>t) \pi_i\notag\\={}&\sum_{\mathsf W_{i}\in\mathscr S}\mathbb P^{\mathsf W_i}(X(n)\notin\mathscr W,\forall n\in\mathbb Z\cap[1,t]) \pi_i\notag\\={}&\sum_{\mathsf W_{i}\in\mathscr S}\sum_{\mathsf W_{m_n}\in\mathscr S\smallsetminus\mathscr W, n\in\mathbb Z\cap[1,t]}\pi_ip_{i m_1} \prod_{n\in\mathbb Z\cap[1,t-1]} p_{{m_{\vphantom{1}n}}{m_{n+1}}}\notag\\={}&\sum_{\mathsf W_{m_n}\in\mathscr S\smallsetminus\mathscr W, n\in\mathbb Z\cap[1,t]}\pi_{m_1}\prod_{n\in\mathbb Z\cap[1,t-1]} p_{{m_{\vphantom{1}n}}{m_{n+1}}},\label{eq:Markov_hitting_rec}
\end{align}}for all $ t\in\mathbb Z_{>1}$.

Consider the abelian semigroup $\Sigma=\mathbb  Z_{\geq0}$. A semi\-character \cite[p.~92, Definition 2.1]{GTM100} $ \rho:\mathbb Z_{\geq0}\longrightarrow\mathbb C$       must assume the following form: $ \rho(0)=1$; $\rho(s)=[\rho(1)]^s$ for  $s\in\mathbb Z_{>0}$. Since the spectral radius of our recurrence matrix is strictly less than  1, we will only concern ourselves with $ \rho_0(s):=\mathbf1_{\{0\}}(s)$ and $ \rho_\lambda(s):=\lambda^s,s\in\mathbb Z_{>0}$ for $ \lambda\in(-1,0)\cup(0,1)$.

 According to the harmonic analysis on semigroups, a bounded function $ f:\mathbb Z_{\geq0}\longrightarrow \mathbb C$ can be represented as a weighted mixture of semi\-characters $ f(s)=\int _{-1<\lambda<1}\rho_{\lambda}(s)\D\mu(\lambda)$ if and only if it is positive definite   \cite[p.~93, Theorem 2.5]{GTM100}:
the inequality \begin{align}
\sum_{r=1}^m\sum_{s=1}^mc_r\overline{c_s}f(t_r+t_s)\geq0\label{eq:semigrp_pos_def}
\end{align}holds for arbitrary positive integers $m\in\mathbb Z_{>0}$, and arbitrarily chosen $m$-dimensional ``vectors'' $(t_1,\dots,t_m)\in\Sigma^m $, $ (c_1,\dots,c_m)\in\mathbb C^m$.

In the following theorem, we  check the positive definiteness criterion \eqref{eq:semigrp_pos_def} for the   hitting time distribution $ 1-H_{\mathscr W}(s+2),s\in\mathbb Z_{\geq0}$ on a Markov chain satisfying the detailed balance condition. This would imply that the return time distribution $ 1-R_{\mathscr W}(s+2),s\in\mathbb Z_{\geq0}$  is also a weighted mixture of semicharacters $ \rho_\lambda(s)$ for $ -1<\lambda<1$, according to the finite difference relation \eqref{eq:H_R_rln} in Theorem~\ref{thm:H_R_rln}.
\begin{theorem}[Positive definite return time distributions]\label{thm:cm_R}  Suppose that a non-void pattern $ \mathscr W\subsetneqq \mathscr S$ on a stationary and ergodic Markov chain  $ \mathscr M=(\varOmega,\mathscr F, \{\mathbb P^{\mathsf W}:\mathsf W\in\mathscr S\},\{X(t):t\in\mathbb Z_{\geq0}\}, \{\widehat{\theta}_n:n\in\mathbb  Z_{\geq0}\})$ satisfies   $ \pi_ip_{ij}=\pi_jp_{ji}$ for all $ \mathsf W_i,\mathsf W_j\in\mathscr S\smallsetminus\mathscr W$.   Then, the cumulative distribution for the return time to $ \mathscr W$ has the following  integral representation\begin{align}
\frac{1-R_{\mathscr W}(s+2)}{1-R_{\mathscr W}(2)}={}&\frac{\sum\limits_{\mathsf W_i\in\mathscr W}\mathbb P^{\mathsf W_i}(\tau_{  \mathscr W}> s+1) \pi_i}{\sum\limits_{\mathsf W_i\in\mathscr W}\mathbb P^{\mathsf W_i}(\tau_{  \mathscr W}> 1) \pi_i}\notag\\={}&\int \rho_\lambda(s)\D \mathscr P_{\mathscr W}(\lambda),\quad s\in\mathbb Z_{\geq0},\label{eq:R_A_pos_def}
\end{align}for a certain probability measure $ \mathscr P_{\mathscr W}(\lambda)$ supported on $ \lambda\in(-1,1)$.

 \end{theorem}\begin{IEEEproof}   By \eqref{eq:Markov_hitting_rec}, we have {\allowdisplaybreaks\begin{align}&
\sum_{\mathsf W_{i}\in\mathscr S}\mathbb P^{\mathsf W_i}(\tau_{\mathscr W}>r+s+1) \pi_i\notag\\={}&\sum_{\mathsf W_{m_n}\in\mathscr S\smallsetminus\mathscr W, n\in\mathbb Z\cap[1,r+s+1]}\pi_{m_1}\prod_{n\in\mathbb Z\cap[1,r+s]} p_{{m_{\vphantom{1}n}}{m_{n+1}}}\notag\\={}&\sum_{\mathsf W_{m_n}\in\mathscr S\smallsetminus\mathscr W, n\in\mathbb Z\cap[1,r+s+1]}\pi_{m_1}\times\notag\\{}&\times\prod_{n\in\mathbb Z\cap[1,r]} p_{{m_{\vphantom{1}n}}{m_{n+1}}}\prod_{n\in\mathbb Z\cap[r+1,r+s]} p_{{m_{\vphantom{1}n}}{m_{n+1}}},
\end{align}}for $ r,s\in\mathbb Z_{>0}$. We then apply the identity   $ \pi_ip_{ij}=\pi_jp_{ji},\forall \mathsf W_i,\mathsf W_j\in\mathscr S\smallsetminus\mathscr W$ to the product over  $ n\in\mathbb Z\cap[1,r]$, which brings us {\allowdisplaybreaks\begin{align}&
\sum_{\mathsf W_{i}\in\mathscr S}\mathbb P^{\mathsf W_i}(\tau_{\mathscr W}>r+s+1) \pi_i\notag\\={}&\sum_{\mathsf W_{m_n}\in\mathscr S\smallsetminus\mathscr W, n\in\mathbb Z\cap[1,r+s+1]}\pi_{m_1}\times\notag\\{}&\times\prod_{n\in\mathbb Z\cap[1,r]} \frac{\pi_{m_{n+1}}}{\pi_{m_{n}}}p_{{}{m_{n+1}}m_{\vphantom{1}n}}\prod_{n\in\mathbb Z\cap[r+1,r+s]} p_{{m_{\vphantom{1}n}}{m_{n+1}}}\notag\\={}&\sum_{\mathsf W_{m_{r+1}}\in\mathscr S\smallsetminus\mathscr W}\mathbb P^{\mathsf W_{m_{r+1}}}(\tau_{\mathscr W}> r)\mathbb P^{\mathsf W_{m_{r+1}}}(\tau_{\mathscr W}>s)\pi_{m_{r+1}}
\end{align}}for $ r,s\in\mathbb Z_{>0}$. In other words, we have verified \begin{align}&
\sum_{\mathsf W_{i}\in\mathscr S}\mathbb P^{\mathsf W_i}(\tau_{\mathscr W}>r+s+1) \pi_i\notag\\={}&\sum_{\mathsf W_{i}\in\mathscr S\smallsetminus\mathscr W}\mathbb E^{\mathsf W_{i}}(\mathbf1_{(\tau_{\mathscr W}> r)})\mathbb E^{\mathsf W_{i}}(\mathbf1_{(\tau_{\mathscr W}> s)})\pi_{i}
\end{align} for $r,s\in\mathbb Z_{>0}$. Using the Markov property, one can readily generalize the identity above to $ r,s\in\mathbb Z_{\geq0}$.  Thus, we have \begin{align}&
\sum_{r=1}^m\sum_{s=1}^m\sum_{\mathsf W_{i}\in\mathscr S}c_r\overline{c_s}\mathbb P^{\mathsf W_i}(\tau_{\mathscr W}>t_{r}+t_{s}+1) \pi_i\notag\\={}&\sum_{\mathsf W_{i}\in\mathscr S\smallsetminus\mathscr W}\left\vert\mathbb E^{\mathsf W_{i}}\left(\sum_{r=1}^{m}c_r\mathbf1_{(\tau_{\mathscr W}> t_{r})}\right)\right\vert^{2}\pi_{i}\geq0
\end{align}for  $ t_r,t_s\in\mathbb Z_{\geq0}$. By the positive definiteness criterion in   \eqref{eq:semigrp_pos_def}, we see that $ 1-H_{\mathscr W}(s+2),s\in\mathbb Z_{\geq0}$ is a weighted mixture of semi\-characters. In view of the recursion identity relating   hitting and return time distributions \eqref{eq:H_R_rec}, we have confirmed the statement in \eqref{eq:R_A_pos_def}.
\end{IEEEproof}

With bin sizes  far larger than 2 time units on the Markov chain, our $ L_{ii}$ is nearly a continuous random variable and all the period-2 oscillatory decays $ \rho_\lambda(s)$ for $ -1<\lambda<0$ behave just like noise terms in the histogram. Then, Theorem~\ref{thm:cm_R} tells us that the probability distribution for $ L_{ii}$ can be  approximated by  a  weighted  mixture\footnote{We note that there exist exceptions to stationarity and ergodicity in realistic documents. A novel may involve the birth and/or death of leading/supporting characters. An academic treatise may place particular emphasis on certain concepts in specific chapters. In such scenarios, the overall  recurrence kinetics may still follow the generic law given in   \eqref{eq:multi_exp}, as a consequence of both detailed balance (in locally applied Markov models) and a heterogeneous mixture of several stationary and ergodic Markov models that patch together as a whole.
 }  of exponential decays $ e^{-kL_{ii}}$, in the continuum limit.\subsection{A statistical criterion for numerical independence between different word patterns\label{app:sigm}
}If we define $\widetilde L_{ij} $ as the effective length of a text fragment free from word  pattern $\mathsf W_j$ that is simultaneously flanked by $\mathsf W_i$ to the left and by $\mathsf W_j$ to the right, then $ \sum_{\mathsf W_i\in\mathscr S}\pi_i\mathbb P(\widetilde L_{ij}>t)\sim \sum_{\mathsf W_i\in\mathscr S}\pi_i\mathbb P^{\mathsf W_i}(\tau_{j}>t)$ represents the hitting time distribution to the Markov state $\mathsf W_j$. According to Theorem~\ref{thm:H_R_rln}, the hitting time distribution can be obtained by integrating the return time distribution $ \mathbb P(\widetilde L_{jj}>t)\sim \mathbb P^{\mathsf W_j}(\tau_j>t)\sim \sum_nc_ne^{-k_nt}$. Here, the positive weights $c_n>0$ are normalized to one: $ \sum_nc_n=1$. In other words, for fixed $\mathsf W_j$, we can predict some statistical properties of $\widetilde L_{ij} $ by analyzing the recurrence statistic $ \widetilde L_{jj}$, as described in the theorem below.
\begin{theorem}[Mean and variance for the logarithm of hitting time]\label{thm:mean_var}Using the continuous time approximation, we have the following relations for a fixed $\mathsf W_j$:\begin{align}\mathbb E \log\widetilde L_{ij}={}&\frac{\langle\widetilde L_{jj}\log \widetilde L_{jj}\rangle}{\langle\widetilde L_{jj}\rangle}-1,\label{eq:mean_conv}\\\mathbb E (\log\widetilde L_{ij}-\mathbb E \log\widetilde L_{ij})^{2}={}&\frac{\langle\widetilde L_{jj}(\ell_*-\log \widetilde L_{jj})^{2}\rangle}{\langle\widetilde L_{jj}\rangle},\label{eq:var_conv}\end{align}where $\ell_*$ equals the right-hand side of \eqref{eq:mean_conv}, and the expectation $ \mathbb E$ denotes  average over all the states $\mathsf W_i $, weighted by $ \pi_i$.\end{theorem}\begin{IEEEproof}In the continuous time approximation, we can assume that  the probability density function for  $ \widetilde L_{jj}$ is  $\sum_nc_nk_{n}e^{-k_nt} $, (where $ c_n,k_n>0$) so that the probability density function for $ \widetilde L_{ij}$ (weighted average over all the states $\mathsf W_i $) is\begin{align}
\frac{ \displaystyle\sum_nc_ne^{-k_nt} }{\displaystyle\sum_nc_n/k_{n} }=\frac{\displaystyle\sum_nc_ne^{-k_nt} }{\langle\widetilde L_{jj}\rangle}.\label{eq:H_distn_L}
\end{align} In view of \eqref{eq:H_distn_L}, we can compute the left-hand side of \eqref{eq:mean_conv} as\begin{align}&
\int_0^\infty\sum_nc_ne^{-k_nt}\log t\frac{\D t}{\langle\widetilde L_{jj}\rangle}=-\sum_n\frac{c_n(\gamma_0+\log k_n)}{k_{n}\langle\widetilde L_{jj}\rangle}.\end{align}
Meanwhile, we  may evaluate the right-hand side of \eqref{eq:mean_conv} as follows:\begin{align}&
\int_0^\infty\displaystyle\sum_nc_nk_{n}e^{-k_nt}t\log t\frac{\D t}{\langle\widetilde L_{jj}\rangle}-1\notag\\={}&-\displaystyle\sum_n\frac{c_n(\gamma_0-1+\log k_n)}{k_{n}\langle\widetilde L_{jj}\rangle}-1.
\end{align} Thus,  \eqref{eq:mean_conv} is confirmed.

In a similar vein, we can argue that the left-hand side of  \eqref{eq:var_conv} equals\begin{align}&
\int_0^\infty\displaystyle\sum_nc_ne^{-k_nt}\log^{2} t\frac{\D t}{\langle\widetilde L_{jj}\rangle}-\ell_*^2\notag\\={}&\displaystyle\sum_n\frac{c_n[6 \log k_{n} (2 \gamma_{0} +\log k_{n})+6 \gamma ^2_{0}+\pi ^2]}{6k_{n}\langle\widetilde L_{jj}\rangle}-\ell_*^{2},\end{align}
while  the right-hand side of  \eqref{eq:var_conv} amounts to {\allowdisplaybreaks\begin{align}&\int_0^\infty\displaystyle\sum_nc_nk_{n}e^{-k_nt}t(\ell_*-\log t)^{2}\frac{\D t}{\langle\widetilde L_{jj}\rangle}\notag\\={}&\int_0^\infty\displaystyle\sum_nc_nk_{n}e^{-k_nt}t\log^{2} t\frac{\D t}{\langle\widetilde L_{jj}\rangle}-2\ell_*-\ell_*^2\notag\\={}&\displaystyle\sum_n\frac{c_n[6 \log k_{n}  (2 \gamma _{0}-2+\log k_{n} )+6 \gamma_{0}  (\gamma_{0} -2)+\pi ^2]}{6k_{n}\langle\widetilde L_{jj}\rangle}\notag\\{}&+\displaystyle\sum_n\frac{2c_n(\gamma_0+\log k_n)}{k_{n}\langle\widetilde L_{jj}\rangle}-\ell_*^2.\end{align}}This completes the proof of  \eqref{eq:var_conv}.      \end{IEEEproof}

The detailed balance condition $ \pi_ip_{ij}=\pi_jp_{ji}$ means that a Markov chain  is reversible, and the reverse chain has the same transition matrix $ \mathbf P$   \cite[\S4.7, p.~203]{Ross1995}.

By the reversibility of our Markov chain, we may extend the results from the theorem above to a dual situation, namely, the statistical properties for  $ L_{ij} $, the effective length of a text fragment free from word $\mathsf W_i$ that is simultaneously flanked by $\mathsf W_i$ to the left and by $\mathsf W_j$ to the right (as considered in Fig.~\ref{fig:NijLij}).

 Trading $ \widetilde L_{ij}$ (resp.~$\widetilde L_{jj}$) in   \eqref{eq:mean_conv}--\eqref{eq:var_conv} for  $  L_{ij}$ (resp.~$ L_{ii}$), we recover \eqref{eq:sigmoid_para}: for a given \textit{i}, we have used  \eqref{eq:mean_conv} as an estimate for  $ \langle \log L_{ij}\rangle$, and have used $ 1/n_{ij}$ times   \eqref{eq:var_conv} as an estimate for the variance of $ \langle \log L_{ij}\rangle $, which is taken over $n_{ij}$ independent samples.
\ifCLASSOPTIONcompsoc
  \section*{Acknowledgments}
\else
  \section*{Acknowledgment}
\fi

We thank N.\ Chomsky and S.\ Pinker for their inputs on several problems of linguistics. We thank X.\ Sun for  discussions on neural networks. We thank X.\ Wan, R.\ Yan and D.\ Zhao for their suggestions on experimental design, during the early stages of this work. We thank two anonymous reviewers, whose thoughtful comments help us improve the presentation of this work.

\ifCLASSOPTIONcaptionsoff
  \newpage
\fi




%

%
\begin{IEEEbiographynophoto}{Weinan E}
 received his B.S. degree in mathematics from the University of Science and Technology of China in 1982, and his Ph.D. degree in mathematics from the University of California, Los Angeles in 1989. He has been a full professor at the Department of Mathematics and the Program in Applied and Computational Mathematics, Princeton University since 1999. Prof.\ E has made significant contributions to numerical analysis,  fluid mechanics, partial differential equations,  multiscale modeling, and stochastic processes.

His monograph \textit{Principles of Multiscale Modeling} (Cambridge University Press, 2011) is a standard reference for mathematical modeling of  physical systems on multiple length scales. He was awarded the Peter Henrici Prize in 2019 for his recent contributions to machine learning.
\end{IEEEbiographynophoto}

\begin{IEEEbiographynophoto}{Yajun Zhou}
received his B.S. degree in Physics from Fudan University
(Shanghai, China) in 2004, and his Ph.D. degree in Chemistry from Harvard University in 2010. Having undertaken postdoctoral training in applied and computational mathematics at Princeton University, he now works in the Laboratory for Natural Language Processing \& Cognitive Intelligence at Beijing Institute of Big Data Research. Dr.\ Zhou has published in numerical analysis, special functions, number theory, electromagnetic theory, quantum theory, probability theory and stochastic processes, solving several open problems in the related fields.

A book chapter in \textit{Elliptic integrals, elliptic functions and modular forms in quantum field theory} (Springer, 2019) surveys  his   proofs of various mathematical conjectures. Dr.\ Zhou is a polyglot, with a working knowledge of 24 languages spanning 7 language families. \end{IEEEbiographynophoto}





\end{document}